%% file: main.tex
\documentclass{article}

\usepackage[main, final]{neurips_2025}

\usepackage[utf8]{inputenc} 
\usepackage[T1]{fontenc}    
\usepackage{hyperref}       
\usepackage{url}            
\usepackage{booktabs}       
\usepackage{amsfonts}       
\usepackage{nicefrac}       
\usepackage{microtype}      
\usepackage{xcolor}         
\usepackage{graphicx}    
\usepackage{subcaption}  

\usepackage{amsmath}
\usepackage{amssymb}
\usepackage{mathtools}
\usepackage{amsthm}

\usepackage{algorithm}
\usepackage{algpseudocode}
\usepackage{wrapfig}
\usepackage{enumitem}
\usepackage{setspace}

\usepackage{multirow}
\usepackage{graphicx}
\usepackage[normalem]{ulem}
\useunder{\uline}{\ul}{}

\makeatletter
\newcommand{\ie}{i.e.\@ifnextchar.{\!\@gobble}{}}
\newcommand{\eg}{e.g.\@ifnextchar.{\!\@gobble}{}}
\newcommand{\etc}{etc\@ifnextchar.{}{.\@}}
\makeatother

\renewcommand{\citet}{\citep}

\theoremstyle{plain}
\newtheorem{theorem}{Theorem}[section]
\newtheorem*{theorem*}{Theorem}

\newtheorem{lemma}[theorem]{Lemma}
\newtheorem*{lemma*}{Lemma}
\newtheorem{corollary}[theorem]{Corollary}
\theoremstyle{definition}
\newtheorem{definition}[theorem]{Definition}

\theoremstyle{remark}

\title{Enhancing Graph Classification Robustness with Singular Pooling}

\author{%
  Sofiane Ennadir\thanks{Equal contribution.} \\
  King AI Labs,  Microsoft Gaming \\
  \texttt{sofiane.ennadir@king.com} \\
  \And
  Oleg Smirnov\footnotemark[1] \\
  King AI Labs, Microsoft Gaming \\
  \texttt{oleg.smirnov@microsoft.com} \\
  \And
  Yassine Abbahaddou \\
  LIX, École Polytechnique, IP Paris \\
\texttt{yassine.abbahaddou@polytechnique.edu} \\
  \And
  Lele Cao \\
  King AI Labs, Microsoft Gaming \\
  \texttt{lelecao@microsoft.com} \\
  \And
  Johannes F. Lutzeyer \\
  LIX, École Polytechnique, IP Paris\\
\texttt{johannes.lutzeyer@polytechnique.edu} \\
}

\begin{document}

\maketitle

\begin{abstract}
  \input{main_paper/0_Abstract}
\end{abstract}

\section{Introduction}\label{sec:introduction}
\input{main_paper/1_introduction}

\section{Related Work}\label{sec:related_work}
\input{main_paper/2_related_work}

\section{Preliminaries}\label{sec:preliminaries}
\input{main_paper/3_preliminaries}

\section{On the Robustness of Pooling Operations}\label{sec:robustness_analysis}
\input{main_paper/4_robustness_study}

\section{Robust Pooling Through Singular Decomposition}\label{sec:proposed_pooling}
\input{main_paper/5_proposed_pooling}

\section{Experimental Results}\label{sec:exp_results}
\input{main_paper/6_experimental_results}

\section{Conclusion}\label{sec:conclusion}
\input{main_paper/7_Conclusion}

\bibliographystyle{plain}
\bibliography{references}

\newpage

\setcounter{page}{1}

\appendix

\vbox{%
\hsize\textwidth
\linewidth\hsize
\vskip 0.1in
\centering
{\LARGE\bf Supplementary Material: Enhancing Graph Classification Robustness with Singular Pooling\par}
\vspace{2\baselineskip}
}

\section{Proof of Theorem~\ref{theo:main_result}}\label{appendix:theo_gcn}
\input{Appendix/1_theo_gcn}

\section{Proof of Theorem~\ref{theo:result_GIN}}\label{appendix:theo_gin}
\input{Appendix/2_theo_gin}

\section{Proof of Theorem~\ref{theo:robustness_of_rpool}}\label{appendix:theo_r_pool}
\input{Appendix/3_theo_r_pool}

\section{Proof of Lemma~\ref{lemma:permutation_invariance}}\label{appendix:proof_remark}
\input{Appendix/4_proof_remark}

\section{Additional Results}\label{appendix:additional Results}
\input{Appendix/5_Additional_results}

\section{Experimental Details}\label{appendix:experimental_details}
\input{Appendix/6_Experimental_Details}

\newpage
\section*{NeurIPS Paper Checklist}
\input{main_paper/check_list}

\end{document}

%% file: main_paper/0_Abstract.tex
Graph Neural Networks (GNNs) have achieved strong performance across a range of graph representation learning tasks, yet their adversarial robustness in graph classification remains underexplored compared to node classification. While most existing defenses focus on the message-passing component, this work investigates the overlooked role of pooling operations in shaping robustness. We present a theoretical analysis of standard flat pooling methods (sum, average and max), deriving upper bounds on their adversarial risk and identifying their vulnerabilities under different attack scenarios and graph structures. Motivated by these insights, we propose \textit{Robust Singular Pooling (RS-Pool)}, a novel pooling strategy that leverages the dominant singular vector of the node embedding matrix to construct a robust graph-level representation. We theoretically investigate the robustness of RS-Pool and interpret the resulting bound leading to improved understanding of our proposed pooling operator. While our analysis centers on Graph Convolutional Networks (GCNs), RS-Pool is model-agnostic and can be implemented efficiently via power iteration. Empirical results on real-world benchmarks show that RS-Pool provides better robustness than the considered pooling methods when subject to state-of-the-art adversarial attacks while maintaining competitive clean accuracy. Our code is publicly available at:
\href{https://github.com/king/rs-pool}{https://github.com/king/rs-pool}.

%% file: main_paper/1_introduction.tex
Graph Neural Networks (GNNs)~\citep{Kipf:2017tc, xu2019powerful, Velickovic:2018we} have demonstrated strong performance on graph-structured data and are now a standard tool for learning node and graph representations. However, recent studies reveal that GNNs are susceptible to adversarial perturbations -- small, deliberately designed changes to the adjacency matrix or node features that can drastically affect predictions~\citep{gunnemann2022graph}. This vulnerability raises concerns about their reliability and consequently, enhancing the adversarial robustness of GNNs has become a key research focus. Efforts include the design of novel attack methods~\citep{dai2018adversarial, zugner2018adversarial, zugner2019adversarial, geisler2021robustness} and corresponding defense mechanisms~\citep{gnn_jaccard, gnn_guard, gosch2024adversarial}. However, prior work primarily investigates node-classification tasks, with limited attention paid to adversarial robustness in graph-classification settings~\citep{review_attacks}. Yet, robustness at the graph level is equally critical, particularly in domains where entire graphs, rather than individual nodes, constitute the prediction target. In drug discovery, for example, molecules are modeled as graphs, and predicting properties such as efficacy or toxicity is a graph classification problem~\citep{qabel2022structure, kearnes2016molecular}. Adversarial perturbations, editing a bond type for instance, can lead to erroneous evaluations with serious implications in precision medicine. Similarly, in cybersecurity, classifying control-flow graphs of binary programs aids in malware detection~\citep{peng2025evading}; subtle perturbations could allow malicious code to evade the anomaly classification.

Existing adversarial defenses for GNNs primarily focus on the message-passing component, leveraging techniques such as attention mechanisms~\citep{gnn_guard}, weight regularization~\citep{abbahaddou2024bounding, ennadir2024simple}, and adjacency preprocessing~\citep{gnn_jaccard}. While these approaches are applicable to both node and graph classification tasks, graph-level prediction introduces additional challenges that go beyond message passing~\citep{gilmer2017}, particularly in constructing global representations from structurally diverse inputs. This step relies on pooling mechanisms~\citep{cai2021graph, Duvenaud_flat_pooling}, which aggregate node features into fixed-size graph embeddings. Pooling operations are generally categorized into two types~\citep{liu2022graph}: \textit{hierarchical pooling}, which performs multi-stage graph coarsening to produce progressively smaller subgraphs~\citep{cai2021graph}, and \textit{flat pooling} (or readout functions), which compute a global representation in a single step using aggregation functions such as sum, average, or max~\citep{Duvenaud_flat_pooling}. While hierarchical pooling is a general architectural tool used throughout the network, flat pooling specifically produces the final representation used in graph-level classification. Given the importance of this final aggregation step, it is critical to investigate how flat pooling functions influence the robustness of GNNs.

This work offers a complementary perspective on GNN robustness by focusing on the pooling stage, which plays a critical role in bridging node-level representations and graph-level embeddings. We center our analysis on flat pooling methods, which are widely used for their simplicity, efficiency, and strong empirical performance. Through theoretical analysis, we examine how different pooling strategies influence the expected robustness and provide insights on selecting pooling functions that balance robustness and performance under varying attack scenarios.

To pursue this goal, we begin by formalizing adversarial attacks in the context of graph classification. This framework establishes an upper bound on the model's expected adversarial risk, allowing us to quantify GNN vulnerability within defined neighborhoods of input graphs. We then theoretically analyze common pooling operations, derive their respective robustness bounds, and provide insights into their behavior under perturbations. These bounds reflect both graph structure and model-specific factors such as message-passing weights. Motivated by this analysis, we propose \textit{Robust Singular Pooling (RS-Pool)}, a novel pooling mechanism that constructs graph-level representations using the dominant right-singular vector of the node embedding matrix. RS-Pool improves robustness by filtering out noise-sensitive components. RS-Pool is fully differentiable, model-agnostic, and compatible with existing GNN architectures and defense methods. 
Since, we rely solely on the dominant singular vector in our pooling operation, we do not require the expensive computation of the entire singular value decomposition and can instead employ the far more efficient power iteration method enabling practical deployment. 
Empirical results across multiple adversarial settings confirm that RS-Pool achieves strong robustness with minimal impact on clean accuracy and runtime performance.

The main contributions of this work are threefold. First, we introduce a theoretical framework that quantifies the effect of flat pooling operations on the resulting robustness of GNNs in graph classification. Second, we propose \textit{RS-Pool}, a novel pooling mechanism that leverages the dominant singular vector of the node representation matrix to improve resilience to adversarial perturbations. Third, we develop an efficient approximation algorithm for RS-Pool and empirically validate its effectiveness in enhancing model robustness under adversarial attacks.

%% file: main_paper/2_related_work.tex
The adversarial robustness of graph neural networks (GNNs) has attracted substantial attention~\cite{gunnemann2022graph,sun2022adversarial, kummer2024on, ennadir2023unboundattack}, spurring a variety of attack methodologies. These attacks are typically formulated as optimization problems that seek minimal perturbations of the adjacency matrix or node features to alter model predictions. They fall into two main categories: evasion attacks, which alter the graph after training~\citep{xu2019topology,zugner2018adversarial,geisler2021robustness,gosch2024adversarial}, and poisoning attacks, which introduce modifications during training~\citep{zugner2020adversarial,lingam2024rethinking}. Although most studies target node classification, techniques such as Projected Gradient Descent (PGD) also extend to graph classification. Nevertheless, only a few methods explicitly address the graph‐level setting. For example, \citet{wan_bayesian} leverage Bayesian optimization to mount attacks on graph classifiers, while \citet{dai2018adversarial} evaluate reinforcement learning, gradient‐based optimization, and genetic algorithms that iteratively refine graphs via a fitness‐driven evolutionary process.

On the defense side, numerous strategies have been proposed to strengthen GNN robustness such as input preprocessing \cite{gnnsvd}. For instance, \citet{gnn_jaccard} remove potentially adversarial edges using Jaccard similarity. These techniques aim to suppress perturbations before message passing. More recent approaches intervene in the message-passing process itself. GNNGuard~\citep{gnn_guard} employs attention-based edge pruning to discard vulnerable connections, and SoftMedian~\citep{geisler2021robustness} introduces a robust aggregation scheme that down-weights neighbors based on their distance to the dimension-wise median, mitigating the impact of outliers or malicious inputs. Other architectural defenses include GCORN~\citep{abbahaddou2024bounding}, which enforces orthogonal weights to enhance stability. Beyond architectural changes, adversarial training remains highly effective. For example, \citet{gosch2024adversarial} train GNNs on perturbed graphs, increasing robustness to future attacks. Complementing empirical defenses, certified methods are gaining traction. Techniques such as randomized smoothing~\citep{bojchevski_certificate_2020}, building on earlier work~\citep{zugner_2019,bojchevski_2019}, offer probabilistic guarantees that predictions remain stable under bounded perturbations.

In parallel, research on pooling strategies has also made great strides \cite{zhang2018end, ma2020path, lee2019self, bianchi2020spectral}, where different pooling methods have been presented to enhance the model's predictive ability. In this work, we aim to connect these two research directions, by examining how flat pooling operations affect the expected robustness. Rather than competing with existing defenses, we offer a complementary viewpoint by analyzing the final pooling stage -- a crucial but often overlooked component in robustness research for graph-level prediction. To the best of our knowledge, this is the first systematic study of pooling in the context of adversarial defense for graph classification. While recent work by~\citet{wang2024comprehensive} offers a broad empirical benchmark of hierarchical pooling methods, assessing their robustness and generalization, their study centers on graph coarsening and does not address flat pooling or its theoretical implications in adversarial settings.

%% file: main_paper/3_preliminaries.tex
In this section, we introduce the fundamental concepts and notation that underpin our work.

\textbf{Notation and Setup.}
Let $G = (V, E)$ be a graph, where $V$ and $E$ denote its sets of nodes and edges, respectively. We use $n = |V|$ and $m = |E|$ to represent the number of nodes and edges. The neighborhood of a node $v \in V$ is denoted by $\mathcal{N}(v) = \{ u \colon (v, u) \in E \}$. The degree of node $v$ is thus $\deg(u) = |\mathcal{N}(v)|$. A graph is typically represented by its adjacency matrix $A \in \mathbb{R}^{n \times n}$ and, when available, a node feature matrix $X \in \mathbb{R}^{n \times d}$, where $d$ is the feature dimensionality.

\textbf{Message Passing GNNs.}
A GNN consists of a sequence of neighborhood aggregation layers that update node representations by incorporating information from local neighborhoods. Let $h_v^{(0)}$ denote the initial feature vector of node $v$. The hidden representation of node $v$ at layer $\ell \leq L$ is updated as:
\begin{equation*}
    \begin{split}
        a_v^{(\ell)} &= \textrm{AGGREGATE}^{(\ell)} \Big( \big\{ h_u^{(\ell-1)} \colon u \in \mathcal{N}(v) \big\} \Big); \quad
        h_v^{(\ell)} = \textrm{COMBINE}^{(\ell)} \Big(h_v^{(\ell-1)}, a_v^{(\ell)} \Big),
    \end{split}
\end{equation*}
where $\textrm{AGGREGATE}^{(\ell)}$ is a permutation-invariant function that summarizes the features of $v$’s neighbors, and $\textrm{COMBINE}^{(\ell)}$ integrates this summary with $v$’s previous representation.

\textbf{Pooling Operation.} After $L$ layers of message passing, a permutation-invariant pooling function is applied to the final node representations to obtain a graph-level embedding:
\begin{equation*}
    h_G = \textrm{POOL} \Big( \big\{h_v^{(L)} \colon v \in V \big\} \Big).
\end{equation*}
This step, often referred to as readout or flat pooling, is essential for producing a fixed-size representation of the entire graph. In this study, we aim to understand how this operation affects the expected robustness of GNNs. For tractability, we focus on flat pooling methods that aggregate node representations in a single step, specifically Sum, Average, and Max pooling.

%% file: main_paper/4_robustness_study.tex
We start by discussing the concept of the expected robustness in graph classification tasks, followed by a theoretical analysis of widely used flat pooling operations. 

\subsection{Expected Graph Robustness}
We consider a set of input graphs represented by their adjacency matrices and node features with associated labels, denoted as ${ (A_1, X_1, y_1), \ldots, (A_N, X_N, y_N) } \in (\mathcal{A}, \mathcal{X}, \mathcal{Y})^N$, sampled from an underlying distribution $\mathcal{D}$ over $(\mathcal{A}, \mathcal{X}, \mathcal{Y})$. The goal of graph classification is to learn a function $f \colon (\mathcal{A}, \mathcal{X}) \rightarrow \mathcal{Y}$ that predicts a target label by minimizing the expected classification risk under $\mathcal{D}$.

In this work, we focus on the black-box evasion setting, where the attacker has no access to the trained model $f$ or its training data and cannot modify them. For the purpose of analysis, we extend the framework introduced in~\citet{ennadir2024simple} to the graph classification context. Let $(A, X) \in (\mathcal{A}, \mathcal{X})$ be an input graph with label $y \in \mathcal{Y}$. We assess the model’s robustness by examining its behavior in the neighborhood of the input graph, measuring the expected risk of prediction changes under bounded perturbations. Given a perturbation budget $\epsilon$, the expected adversarial risk of $f$ with respect to $\mathcal{D}$ is defined as:
\begin{equation}\label{equation:robustness_definition}
\mathcal{R}_{\epsilon}[f] = \mathop{\mathbb{E}}_{\substack{(A, X) \sim \mathcal{D} \\ (\tilde{A}, \tilde{X}) \in \mathcal{N}_{\epsilon}(A, X)}} [d_{\mathcal{Y}}(f(\tilde{A}, \tilde{X}), f(A, X))],
\end{equation}
where $d_{\mathcal{Y}} = \lVert \cdot \rVert_2$ denotes distance in the output space $\mathcal{Y}$. The set $\mathcal{N}_{\epsilon}(A, X) = \{(\tilde{A}, \tilde{X}) : d_{\mathcal{A}, \mathcal{X}}((A, X), (\tilde{A}, \tilde{X})) < \epsilon\}$ defines the neighborhood of admissible perturbations under the budget $\epsilon$, measured by the input space distance $d_{\mathcal{A}, \mathcal{X}}$. In our work, we consider a distance that reflect both the topology and the node features while taking into account the set of permutation matrices $\Pi$, which can be written as: $d_{\mathcal{A}, \mathcal{X}}((A, X), (\tilde{A}, \tilde{X})) = \min_{\substack{P \in \Pi}} \left\{ \lVert A - P \tilde{A} P^T \rVert_2 + \lVert X - P \tilde{X} \rVert_2 \right\}.$

From a defense perspective, the goal is to minimize the expected risk $\mathcal{R}_{\epsilon}[f]$, thereby ensuring the model's predictions remain stable within the allowed perturbation budget. Bounding this risk offers valuable insights into a model's expected robustness and its sensitivity to adversarial inputs. Based on this, we formalize the notion of GNN robustness in Definition~\ref{def:robustness}.

\begin{definition}
\label{def:robustness}
\textbf{(Expected Robustness)} A graph-based function $f \colon (\mathcal{A}, \mathcal{X}) \rightarrow \mathcal{Y}$ is said to be $(\epsilon, \gamma)$-robust if its expected risk satisfies $\mathcal{R}_{\epsilon}[f] \leq \gamma$. 
\end{definition}

We note that this definition is different from the usually used adversarial robustness in the literature, and rather adopts an \emph{average case} perspective by evaluating robustness over the entire neighborhood $\mathcal{N}_{\epsilon}(A, X)$, in contrast to \emph{worst case} analysis, which considers only the most damaging perturbation. We have chosen to operate within this direction, since we consider that worst case attacks represent a subset of the neighborhood and therefore it follows that upper-bounding this quantity for the average would also give me some insights on the robustness under the worst case criterion. This distinction between average and worst case robustness has been discussed in the literature~\cite{rice2021robustness}.

\subsection{On the Expected Robustness of Flat Pooling Operations}\label{sec:theo_results}

Having established a formal definition of the expected robustness in the context of graph classification, we now apply this framework to assess the robustness properties of commonly used pooling operations. Specifically, we examine two representative models from the message-passing family introduced in Section~\ref{sec:preliminaries}: Graph Convolutional Networks (GCNs)~\citep{Kipf:2017tc} and Graph Isomorphism Networks (GINs)~\citep{xu2019powerful}, both of which are widely adopted in graph-level tasks.

Following message passing, node embeddings are aggregated through a pooling operation to produce a global graph representation suitable for downstream classification. However, this step can introduce information loss, prompting the development of various pooling strategies. In this work, we focus on three widely used flat pooling techniques, which generate graph-level representations through a single aggregation step over node embeddings: Sum, Average, and Max pooling~\citep{Duvenaud_flat_pooling}. While our theoretical analysis centers on these operations, the proposed robustness framework is general and can be extended to other mechanisms. The primary objective is to quantify the influence of each pooling strategy on the expected robustness of the model, as defined in Definition~\ref{def:robustness}. Throughout the remainder of this paper, $\lVert \cdot \rVert$ denotes the operator norm.

\begin{theorem}\label{theo:main_result}
Let $f \colon (\mathcal{A}, \mathcal{X}) \rightarrow \mathcal{Y}$ denote a graph-based function composed of $L$ GCN layers, where the weight matrix of the $\ell$-th layer is denoted by $W^{(\ell)}.$ Under a feature-based adversarial attack with perturbation budget $\epsilon$, the robustness of $f$ in the sense of Definition~\ref{def:robustness} satisfies the following:
\begin{itemize}
    \item If $f$ is based on Max pooling, then it is $(\epsilon, \gamma)$-robust with:
    \begin{center}
        $\gamma = \sqrt{\min\{n,d_L\}} \left (\prod_{\ell=1}^L \lVert W^{(\ell)}\lVert \right )  \max_{u \in V}  \hat{w}_u \epsilon.$  
    \end{center}

    \item If $f$ is based on Sum pooling, then it is $(\epsilon, \gamma)$-robust with:
    \begin{center}
        $\gamma = \left (\prod_{\ell=1}^L \lVert W^{(\ell)}\lVert \right ) \sum_{u \in V}  \hat{w}_u \epsilon.$  
    \end{center}

    \item If $f$ is based on Average pooling, then it is $(\epsilon, \gamma)$-robust with:
    \begin{center}
        $\gamma = \frac{1}{n} \left (\prod_{\ell=1}^L \lVert W^{(\ell)}\lVert \right ) \sum_{u \in V}  \hat{w}_u \epsilon.$  
    \end{center}
    
\end{itemize}
Here, $\hat{w}_u$ denotes the sum of normalized walks of length $L{-}1$ originating from node $u$, and $d_L$ is the output dimensionality of the final layer.
\end{theorem}

Theorem~\ref{theo:main_result} establishes upper bounds on the expected risk for different flat pooling operations. A common factor across all bounds is the product of the weight matrix norms, reflecting the influence of message-passing layers. The second term in each bound differentiates the pooling methods. For Sum pooling, the bound scales with the total normalized walk count, which increases with graph density. Consequently, adversarial noise propagates through multiple paths, weakening robustness and highlighting Sum pooling's vulnerability in dense graphs. In contrast, Average pooling normalizes this term by the number of nodes, diminishing the impact of increased walk counts. As graphs grow larger or sparser, this normalization tightens the bound, offering improved robustness over Sum pooling. 
We observe the bound on Max pooling to be governed by a single node in the graph for which the normalized sum of walks of length $L-1$ is maximal. Consequently, the max-pooling operator more localized than the Sum and Mean Operators for which our bounds depend on global characteristics. 

The robustness of each pooling method further depends on the attack strategy. Under targeted attacks, where high-degree or influential nodes are perturbed, Max pooling can be particularly vulnerable, especially in gradient-based settings that expose critical nodes. However, when less influential nodes are targeted, it may retain robustness. In non-targeted attacks that perturb nodes more uniformly, Max pooling may outperform Sum and Average pooling, as it aggregates over only one node. Overall, the analysis reveals that no single pooling method is optimal across all scenarios. Robustness varies with graph structure and attack type, underscoring the importance of selecting pooling strategies based on the anticipated threat model. Although our analysis focuses on GCNs, the same theoretical framework naturally extends to Graph Isomorphism Networks (GINs).

\begin{theorem}\label{theo:result_GIN}
Let $f \colon (\mathcal{A}, \mathcal{X}) \rightarrow \mathcal{Y}$ be composed of $L$ GIN-layers (with its internal parameter $\zeta = 0$) and let $W^{(\ell)}$ denote the weight matrix of the $\ell$-th MLP layer. We consider the input node feature space to be bounded, \ie, $\lVert X \rVert_2 < B$ for some $B\in \mathbb{R}.$ Under a feature-based adversarial attack with perturbation budget $\epsilon$, the model $f$ is $(\epsilon, \gamma)$-robust with respect to Definition~\ref{def:robustness}, with:

\begin{itemize}[leftmargin=4em]
    \item If $f$ is based on Max pooling: $\gamma = \prod_{\ell=1}^L \lVert W^{(\ell)} \rVert \left( B  L  \left(\max_{u \in V} \deg(u)\right) + \epsilon \right).$

    \item If $f$ is based on Sum pooling: $\gamma = \prod_{\ell=1}^L \lVert W^{(\ell)} \lVert \left( 2B L  |E| + n \epsilon \right).$

    \item If $f$ is based on Average pooling: $\gamma = \sqrt{n, d_{L}} \big(\prod_{\ell=1}^L \lVert W^{(\ell)} \lVert \big) \left( \frac{2B  L |E|}{n} + \epsilon \right).$
\end{itemize}
Here, $|E|$ is the number of edges, $n$ number of nodes and $d_L$ the dimensionality of the final layer $L$.
\end{theorem}

As in the GCN case, the derived upper bounds for GIN highlight the influence of pooling operations on expected robustness. For Sum pooling, the expected adversarial risk grows with both the number of nodes and edges, indicating increased vulnerability in larger or denser graphs. In comparison, Average pooling normalizes by the number of nodes, which reduces the relative impact of individual perturbations and enhances robustness in large or sparse settings. Max pooling produces a bound governed by the maximum node degree, suggesting heightened sensitivity in graphs with hub-like structures, particularly when high-degree nodes are targeted. These findings are consistent with GCN-based analysis and further emphasize that pooling is a key factor in shaping the robustness of GNNs.

%% file: main_paper/5_proposed_pooling.tex
Section~\ref{sec:theo_results} demonstrated that pooling significantly influences the robustness of graph classification models. This aligns with the intuition behind message passing, where node-level perturbations propagate through local neighborhoods and ultimately affect the global representation, depending on the pooling strategy. Influential nodes with high centrality or strong connectivity tend to amplify adversarial effects, spreading their influence and creating compound vulnerabilities. Similar trends have been observed in node classification~\citep{li2023revisiting}, where gradient-based attacks often target structurally important nodes. Mitigating the influence of such nodes can therefore improve robustness.

To address adversarial vulnerabilities, prior work has proposed various preprocessing techniques~\citep{gnn_jaccard,gnnsvd,gnn_guard,geisler2021robustness}. However, these defenses typically operate before or during the message-passing stage and are primarily tailored for node classification tasks, where graphs are large and removing a subset of nodes has limited impact on performance. In contrast, graph classification often involves smaller graphs for which the removal of nodes can be more critical. Removing or down-weighting certain nodes can disrupt information propagation and degrade clean accuracy, which is especially problematic given that it is not known in advance whether a graph has been attacked. As a result, achieving a favorable trade-off between robustness to adversarial inputs and strong performance on clean data remains a key challenge in graph-level learning.

Given these considerations, to avoid the consequences of node removal, we propose an alternative pooling strategy that effectively balances robustness and clean performance. We introduce \textit{Robust Singular Pooling (RS-Pool)}, a novel method that retains informative signals in clean graphs while mitigating the influence of adversarial perturbations. RS-Pool generates the graph-level representation by extracting the dominant right singular vector of the node embeddings. This design is motivated by classical results in matrix perturbation theory~\citep{DavisKahan1970, wedin1972perturbation}, which demonstrate that leading singular vectors are stable under bounded perturbations and encode the most reliable directions of variation.

In the context of graph classification, adversarial attacks distort node embeddings produced by message-passing layers. These distortions are typically localized in nature, due to the limited depth of GNNs (empirically often $L\in \{2,3,4, 5\}$ is observed to be highest-performing \cite{errica2020fair}) and constrained perturbation budgets that restrict modifications to a small subset of nodes~\citep{li2023revisiting}. As a result, the dominant singular vector often remains stable, while less-dominant components are more susceptible to noise. By projecting onto this robust direction, RS-Pool generates a stable and expressive graph representation that maintains performance under adversarial conditions. Formally, let $H \in \mathbb{R}^{n \times d}$ denote the matrix of node representations obtained after the message-passing stage (\eg, from a GCN~\citep{Kipf:2017tc}), where $n$ is the number of nodes and $d$ the features (embedding) dimension. 
We consider the singular value decomposition (SVD): $H = U \Sigma V^\top,$
where $U \in \mathbb{R}^{n \times n}$, $\Sigma \in \mathbb{R}^{n \times d}$, and $V \in \mathbb{R}^{d \times d}$. The singular values $\sigma_1 \geq \sigma_2 \geq \dots \geq \sigma_{\min(n,d)} \geq 0$ quantify the contribution of each singular direction. Let $v_1 \in \mathbb{R}^d$ denote the dominant right singular vector, RS-Pool defines the graph-level representation as a scaled version of~$v_1$:
\begin{align*}
    \text{RS-Pool} \colon \quad & \mathbb{R}^{n \times d} \to \mathbb{R}^d \\
    & H \mapsto \tau \, v_1(H),
\end{align*}
where $\tau \in \mathbb{R}_{>0}$ is a scaling factor controlling the magnitude of the output embedding. The dominant singular vector captures the principal direction of variation in the embedding space, effectively summarizing the global structure encoded in $H$. Scaling ensures the pooled vector retains sufficient energy to serve as a meaningful input to downstream classifiers.

Crucially, RS-Pool also offers inherent robustness. Classical results from matrix perturbation theory~\citep{DavisKahan1970, wedin1972perturbation} show that if the leading singular value $\sigma_1(H)$ is well-separated from the next $\sigma_2(H)$ (we use $\sigma_s$ and $\sigma_s(H)$ interchangeably when unambiguous), then the top singular vector $v_1(H)$ is stable under small perturbations to $H$. This separation condition ensures that adversarial distortions, often low-rank and localized, have limited influence on the dominant representation. Within the expected adversarial risk framework introduced in Definition~\ref{def:robustness}, we now formalize the robustness guarantee of RS-Pool.

\begin{theorem}\label{theo:robustness_of_rpool}
Let $f \colon (\mathcal{G}, \mathcal{X}) \rightarrow \mathcal{Y}$ denote a graph function composed of $L$ GCN layers and using our RS-Pool, where the weight matrix of the $i$-th layer is denoted by $W^{(i)}.$ Under a feature-based adversarial attack with perturbation budget $\epsilon$, $f$ is $(\epsilon, \gamma)$-robust in respect to Definition~\ref{def:robustness} with:
\begin{equation}
    \gamma = \frac{\tau \sqrt2 \epsilon}{\sigma_1 - \sigma_2} \Bigl(\prod_{\ell=1}^L \lVert W^{(\ell)}\lVert\Bigr) 
    \sum_{u=1}^n (\hat{w}_u)^2,    
\end{equation}
with $\hat{w}_u$ denoting the sum of normalized walks of length $(L-1)$ starting from node $u,$ and $\sigma_1 \neq \sigma_2$ being the two dominant singular values. 
\end{theorem}

Theorem~\ref{theo:robustness_of_rpool} provides an upper bound on the expected adversarial risk of the proposed RS-Pool method. As with previous bounds, the expression includes shared terms such as the product of message-passing weight norms and the perturbation budget. However, a distinguishing feature of this bound is its dependence on the spectral gap of the clean representation matrix $H$ and the scaling factor $\tau$. Notably, the bound relies solely on the spectral gap of the unperturbed node embeddings, not the perturbed input, reinforcing our earlier claim that RS-Pool mitigates the influence of adversarial noise. Furthermore, the scaling parameter $\tau$ acts as a tunable control for the representation's sensitivity, allowing the bound to be tightened and thereby improving robustness. This tunability is especially beneficial in settings where the message-passing layers exhibit high sensitivity to input perturbations.
\begin{corollary}\label{corollary:effect_of_tau}
Let $f \colon (\mathcal{G}, \mathcal{X}) \rightarrow \mathcal{Y}$ be the graph-based function considered in Theorem~\ref{theo:robustness_of_rpool}. Let $\gamma$ be the upper-bound derived in Theorem~\ref{theo:robustness_of_rpool}, we have that $f$ is $(\epsilon, \gamma')$-robust with:
\begin{equation}
    \gamma' = \min\{\gamma, 2 \tau\}.
\end{equation}
\end{corollary}
Corollary~\ref{corollary:effect_of_tau} highlights that even when the message-passing component is highly vulnerable to adversarial perturbations, as indicated by large weight norms, RS-Pool can still ensure a bounded expected risk by constraining the representation through the scaling parameter $\tau$. This scenario has motivated prior robustness techniques such as orthogonal regularization~\cite{cisse2017parseval}. The result demonstrates that RS-Pool serves as a stabilizing mechanism, contributing to overall robustness even when earlier layers are sensitive to input perturbations. Moreover, RS-Pool retains the essential property of permutation invariance, ensuring its compatibility with standard GNN architectures.
\begin{lemma}\label{lemma:permutation_invariance}
RS-Pool is permutation invariant; that is, for any pair of isomorphic graphs $G$ and $G_{\Pi}$ with its corresponding permutation matrix $\Pi$, we have $f(A, X) = f(A_{\Pi}, X_{\Pi})$.
\end{lemma}

\textbf{Remark.} Theorem \ref{theo:robustness_of_rpool} does not address cases where the largest singular value has multiplicity greater than one. Such cases are rare in practice, as GNNs' inherent smoothing tends to reduce the feature matrix rank and create a non-trivial spectral gap. Nonetheless, Wedin's Theorem, which underlies our proof, extends to the case where the dominant singular value has multiplicity $r$, yielding an additional factor $\sqrt{r}$ (instead of $\sqrt{2}$) in the upper bound $\gamma$, with the relevant spectral gap $\sigma_r - \sigma_{r+1}$.

\begin{wrapfigure}[10]{r}{0.40\textwidth}
\vspace{-15pt}
\begin{minipage}{\linewidth}
\vspace{-6pt}
\begin{algorithm}[H] 
  \caption{RS-Pool Forward Pass}
  \label{alg:rs_pool}
  \footnotesize 
  \begin{algorithmic}[1]
    \Require $H \in \mathbb{R}^{n \times d}$, $\tau \in \mathbb{R}_{>0}$, $K \in \mathbb{N}_{>0}$
    \State $S \gets H^{\top} H$
    \State $v \gets$ random unit vector in $\mathbb{R}^{d}$
    \For{$t = 1$ \textbf{to} $K$}
      \State $v \gets S v$
      \State $v \gets v / \lVert v \rVert_{2}$
    \EndFor
    \State \Return $\tau H v$
  \end{algorithmic}
\end{algorithm}
\end{minipage}
\vspace{-5pt}
\end{wrapfigure}
\textbf{Implementation and Algorithm.} RS-Pool constructs the graph-level representation by extracting the dominant right-singular vector of the node embedding matrix $H$. As only the top singular vector is needed, computing the costly full SVD is unnecessary. Instead, we employ the truncated power iteration method to efficiently estimate the dominant vector. This iterative approach is scalable with respect to both the number of nodes and the embedding dimension. A key advantage of RS-Pool is that it remains fully differentiable, allowing seamless integration into existing GNN pipelines and enabling end-to-end training via standard backpropagation. Algorithm~\ref{alg:rs_pool} outlines the forward pass of a GNN model utilizing RS-Pool.

\textbf{Computational Complexity.} The main computational overhead of RS-Pool stems from approximating the dominant singular vector using power iteration. Each iteration has a complexity of $\mathcal{O}(n \times d)$, where $n$ is the number of nodes and $d$ the embedding dimension.  In practice, we find that only a small number of iterations (e.g., 2–5) is sufficient to obtain a reliable estimate. This efficiency is supported by the convergence properties of power iteration, which is known to converge geometrically at a rate determined by the ratio $\sigma_2(H) / \sigma_1$(H), where $\sigma_1(H)$ and $\sigma_2(H)$ are the top two singular values of $H$. A well-separated spectral gap ensures rapid convergence, which is commonly observed in GNN embeddings due to dominant low-frequency components. As a result, the overall complexity remains $\mathcal{O}(K \times n \times d)$ for a small $K$, making RS-Pool suitable for real-world graph applications. A comparison of its runtime with other pooling methods is provided in Appendix \ref{app:time_analysis}.

We further note that the fast empirical convergence of power iteration aligns with the theoretical robustness bound in Theorem~\ref{theo:robustness_of_rpool}, where the spectral gap $\sigma_1(H) - \sigma_2(H)$ plays a central role. A larger gap not only tightens the bound (lower $\gamma$), indicating improved robustness, but also ensures faster convergence of the power iteration method used to compute RS-Pool. This dual benefit highlights the practical effectiveness of RS-Pool, as favorable spectral properties commonly observed in GNN embeddings lead to both stable representations and efficient computation. We provide empirical validation of this observation in Appendix~\ref{sec:convergence_analysis}.

%% file: main_paper/6_experimental_results.tex
This section aims to validate our theoretical insights on real-world datasets where we aim to investigate the robustness of RS-Pool in comparison to other considered pooling methods. In line with our theoretical study, we focus on the GCN model in this section while we provide results on the GIN in Appendix~\ref{app:results_gin}. We additionally provide results when considering the adversarial defenses as a backbone GNN in Appendix~\ref{app:results_defense}, showcasing therefore the universality of our insights. For all our experiments, we use a 2 layers GCN (resp. GIN) classifier with identical hyper-parameters across all experiments to guarantee a fair comparison. Each experiment is run $10$ times to reduce the effect of randomness. Additional implementation details are provided in Appendix~\ref{appendix:experimental_details}. 

\textbf{Considered Benchmarks.} We evaluate RS-Pool against classical flat pooling methods, including Average, Max, and Sum pooling, as well as several advanced approaches. We include Self-Attention Graph pooling (SAG)~\cite{lee2019self}, which uses node features and graph structure to compute attention scores and retain informative nodes, and TopK pooling (TopK-P)~\citep{gao2019graph}, which learns to rank nodes via a projection vector and selects a fixed fraction by score. We also consider Path-Integral-based Pooling (PAN-P)~\cite{ma2020path}, a TopK extension that integrates multi-hop structure into scoring. Lastly, we compare with Sort Pooling (Sort-P)~\cite{zhang2018end}, which orders nodes by feature magnitude and concatenates the top-$k$ sorted nodes.

\textbf{Attacks.} We evaluate robustness under three adversarial attack strategies: \textbf{(i)} \textit{Random Attack}, which randomly adds or removes edges in the input graph. The search is performed over $K$ random perturbations, and the one yielding the worst performance is selected; \textbf{(ii)} \textit{Genetic Attack}~\citep{dai2018adversarial}, which uses evolutionary algorithms to generate adversarial graphs via selection, crossover, and mutation; and \textbf{(iii)} \textit{Gradient-Based Attack (PGD)}~\citep{dai2018adversarial}, which greedily modifies the graph structure by targeting edges with the highest gradient magnitude relative to the model's input. For all attacks, we apply a perturbation budget of $\epsilon = 0.3$, allowing up to 30\% of the edges in each graph to be modified. We additionally considered the Bit-Flip Attacks (BFAs)~\citep{kummer2024on, kummer2025on} in Appendix~\ref{app:bit_flip_attacks}. 

\textbf{Datasets.} We conduct experiments on standard graph classification datasets from the TUDataset benchmark~\citep{morris2020tudataset}, spanning diverse domains. In bioinformatics graphs (PROTEINS, D\&D, ENZYMES), small changes to residue contact links can influence biological property predictions. In molecular graphs (NCI1, ER\_MD), altering a bond may change the predicted molecular function. For social networks (IMDB-B, REDDIT-B), edge perturbations, such as fake user interactions, can flip the predicted graph label. In image-based graphs (MSRC\_9), local edits to image patches can lead to misclassification. For all datasets, we use the public train/validation/test splits when available~\citep{errica2020fair}; otherwise, we adopt the same evaluation protocol and report the specific folds used.

\subsection{Experimental Results}

\begin{figure}
    \centering
    \includegraphics[width=\linewidth]{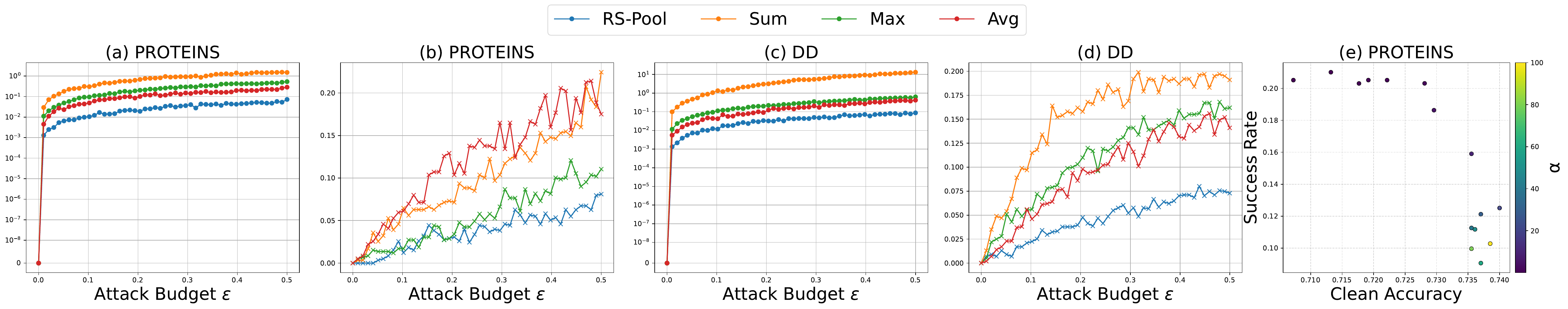}
    \caption{Left: Empirical estimation of adversarial risk $\gamma$ and corresponding attack success rate on PROTEINS ((a), (b)) and D\&D ((c), (d)). Right (e): Study of the effect of the parameter $\alpha$ which inversely controls $\tau$ on the PROTEINS dataset.}
    \label{fig:analysis_of_risk}
\end{figure}

\textbf{Empirical Estimation of Adversarial Risk.} We begin by empirically analyzing the adversarial risk defined in Section~\ref{sec:robustness_analysis}, characterized by the upper bound $\gamma$. To this end, we compute the distance between the pooled representations of clean and adversarially perturbed graphs for each pooling method. Figure~\ref{fig:analysis_of_risk} presents the results of this analysis across the considered pooling strategies. As expected, RS-Pool consistently maintains a smaller distance between clean and perturbed representations. In addition, we see the exact behavior underlined in our theoretical results provided in Theorem~\ref{theo:main_result}, where the Sum is the one with the higher upper-bound, followed by the Max and the Average. This behavior directly correlates with lower attack success rates of our RS-Pool, as shown in subplots (b) and (d). Complete results for all pooling methods are provided in Appendix~\ref{app:additional_results}.

\textbf{On the Effect of the Hyperparameter $\tau$.} As discussed in Section~\ref{sec:proposed_pooling}, the parameter $\tau$ plays a critical role in RS-Pool. While it directly scales the extracted dominant singular vector, its more significant effect lies in modulating the robustness bound in Theorem~\ref{theo:robustness_of_rpool} and influencing the expected risk as detailed in Corollary~\ref{corollary:effect_of_tau}. To validate this theoretically grounded effect, we examine how varying $\tau$ impacts both clean accuracy and robustness. In our experiments, to better isolate the role of the spectral gap, we parameterize $\tau$ as a scaled version of the leading singular value, which is already computed during the power iteration. Specifically, we set $\tau = \sigma_1(X) / \alpha$, where $\alpha > 0$ is a tunable constant. We then study how different values of $\alpha$ affect performance, measuring clean accuracy and attack success rate. As shown in Figure~\ref{fig:analysis_of_risk} (e), increasing $\alpha$ (and thus decreasing $\tau$) leads to a lower attack success rate, indicating improved robustness. However, this comes at the cost of clean accuracy, which peaks at an intermediate $\alpha$ before declining. These results highlight the importance of selecting an appropriate value of $\alpha$ to achieve a desirable trade-off between robustness and predictive performance.

\input{Tables/1_pgd_attack}

\textbf{Results Analysis.} We evaluate the performance of RS-Pool and baseline pooling methods under adversarial attacks. Table~\ref{tab:main_results} presents classification accuracy on clean and perturbed data across datasets and attack types. RS-Pool consistently performs well on clean inputs and significantly outperforms baselines under adversarial conditions. The only exceptions occur in two cases involving the Random attack, where the limited perturbation budget results in low attack success, allowing some baselines to retain higher accuracy due to strong clean performance. Under stronger attacks such as PGD, RS-Pool shows notable gains, achieving up to a 14\% uplift over the second-best method on the D\&D dataset. This advantage is more pronounced on datasets like D\&D, which contain larger and denser graphs where the leading singular vector captures most of the informative signal, enabling more robust representation through RS-Pool. Similar observations hold for GIN and other adversarial defenses, as shown in Appendix~\ref{app:results_gin} and~\ref{app:results_defense}.

\input{Tables/4_results_features}

\textbf{Feature-Based Robustness.}  
To further demonstrate the value of our theoretical analysis, and inline with the considered setup, we evaluate adversarial attacks targeting node features. We adopt the same attack strategies used in the structural setting: \textit{Random}, which adds noise to features at random, and \textit{PGD}, a gradient-based attack. Table~\ref{tab:results_features} reports clean and attacked accuracies (with standard deviations) for our proposed pooling method, RS-Pool, alongside other baseline pooling strategies across multiple datasets. Consistent with the structural attack results, RS-Pool achieves higher attacked accuracy, highlighting its improved robustness against feature perturbations.

%% file: Tables/1_pgd_attack.tex
\begin{table}[]
\small
\caption{Clean and Attacked classification accuracy ($\pm$ standard deviation) of the RS-Pool and other pooling strategies on different graph classification dataset when subject to adversarial attacks.}
\label{tab:main_results}
\renewcommand{\arraystretch}{1.05}
\resizebox{\columnwidth}{!}{%
\begin{tabular}{llcccccccc}
\hline
Dataset                   & Attack  & Sum                     & Average        & Max            & SAG                     & TopK-P         & PAN-P                   & Sort-P                  & RS-Pool                 \\ \hline
\multirow{4}{*}{PROTEINS} & Clean   & 74.2 $\pm$ 3.1          & 70.8 $\pm$ 2.2 & 73.2 $\pm$ 2.3 & 72.3 $\pm$ 3.3          & 73.2 $\pm$ 2.9 & \textbf{74.7 $\pm$ 1.7} & 72.3 $\pm$ 2.3          & 73.5 $\pm$ 2.9          \\
                          & PGD     & 45.8 $\pm$ 2.9          & 34.2 $\pm$ 2.1 & 28.2 $\pm$ 3.5 & 45.5 $\pm$ 2.1          & 48.5 $\pm$ 1.2 & 33.3 $\pm$ 1.5          & 31.8 $\pm$ 2.9          & \textbf{51.9 $\pm$ 3.6} \\
                          & Random  & 68.3 $\pm$ 3.1          & 65.7 $\pm$ 2.3 & 42.9 $\pm$ 4.1 & 61.9 $\pm$ 5.1          & 68.2 $\pm$ 1.8 & 68.5 $\pm$ 2.1          & 64.3 $\pm$ 3.2          & \textbf{70.4 $\pm$ 3.1} \\
                          & Genetic & 66.1 $\pm$ 2.3          & 65.3 $\pm$ 3.2 & 39.8 $\pm$ 5.0 & 62.9 $\pm$ 3.8          & 66.5 $\pm$ 2.2 & 66.4 $\pm$ 2.7          & 63.7 $\pm$ 2.1          & \textbf{68.2 $\pm$ 1.8} \\ \cline{2-10} 
\multirow{4}{*}{D\&D}     & Clean   & 75.1 $\pm$ 0.6          & 70.1 $\pm$ 0.5 & 74.1 $\pm$ 0.6 & 71.3 $\pm$ 0.4          & 74.4 $\pm$ 1.1 & \textbf{75.7 $\pm$ 2.8} & 74.9 $\pm$ 1.7          & 74.6 $\pm$ 0.7          \\
                          & PGD     & 6.7 $\pm$ 2.0           & 12.6 $\pm$ 3.6 & 8.7 $\pm$ 5.2  & 9.0 $\pm$ 3.8           & 5.9 $\pm$ 2.9  & 9.6 $\pm$ 2.8           & 16.6 $\pm$ 1.9          & \textbf{30.4 $\pm$ 3.2} \\
                          & Random  & 61.9 $\pm$ 3.4          & 64.4 $\pm$ 2.9 & 19.7 $\pm$ 2.4 & 62.8 $\pm$ 2.4          & 58.6 $\pm$ 1.1 & 61.9 $\pm$ 2.4          & 50.3 $\pm$ 2.9          & \textbf{66.7 $\pm$ 2.1} \\
                          & Genetic & 60.5 $\pm$ 4.2          & 63.1 $\pm$ 2.6 & 21.4 $\pm$ 1.5 & 62.5 $\pm$ 2.8          & 57.7 $\pm$ 0.6 & 56.1 $\pm$ 3.6          & 57.7 $\pm$ 4.3          & \textbf{67.2 $\pm$ 2.9} \\ \cline{2-10} 
\multirow{4}{*}{NCI1}     & Clean   & \textbf{70.6 $\pm$ 0.8} & 67.9 $\pm$ 1.6 & 68.2 $\pm$ 1.2 & 70.4 $\pm$ 0.9          & 70.9 $\pm$ 0.9 & 70.4 $\pm$ 1.7          & 69.5 $\pm$ 0.4          & 70.1 $\pm$ 1.2          \\
                          & PGD     & 23.4 $\pm$ 1.1          & 14.1 $\pm$ 2.1 & 19.7 $\pm$ 1.7 & 22.3 $\pm$ 1.3          & 23.2 $\pm$ 2.7 & 23.9 $\pm$ 2.3          & 23.5 $\pm$ 1.1          & \textbf{26.3 $\pm$ 1.8} \\
                          & Random  & 26.1 $\pm$ 0.7          & 20.6 $\pm$ 2.6 & 9.6 $\pm$ 5.1  & 19.8 $\pm$ 1.5          & 25.8 $\pm$ 1.8 & 24.7 $\pm$ 2.1          & 24.3 $\pm$ 1.2          & \textbf{27.2 $\pm$ 0.1} \\
                          & Genetic & 25.9 $\pm$ 0.8          & 19.7 $\pm$ 2.7 & 9.3 $\pm$ 2.2  & 18.7 $\pm$ 1.2          & 23.9 $\pm$ 2.4 & 19.1 $\pm$ 2.6          & 25.8 $\pm$ 1.4          & \textbf{26.9 $\pm$ 0.7} \\ \cline{2-10} 
\multirow{4}{*}{ENZYMES}  & Clean   & \textbf{33.4 $\pm$ 4.9} & 27.7 $\pm$ 4.9 & 27.7 $\pm$ 1.6 & 26.1 $\pm$ 0.9          & 27.7 $\pm$ 4.3 & 28.3 $\pm$ 3.6          & 29.7 $\pm$ 2.1          & 32.8 $\pm$ 4.6          \\
                          & PGD     & 2.7 $\pm$ 2.1           & 3.3 $\pm$ 1.3  & 9.8 $\pm$ 2.3  & 8.9 $\pm$ 2.1           & 6.7 $\pm$ 1.4  & 1.7 $\pm$ 1.3           & 5.6 $\pm$ 2.1           & \textbf{11.9 $\pm$ 2.7} \\
                          & Random  & 7.4 $\pm$ 3.4           & 8.3 $\pm$ 2.3  & 2.2 $\pm$ 1.6  & 3.8 $\pm$ 1.6           & 10.6 $\pm$ 1.3 & 5.6 $\pm$ 2.1           & 9.4 $\pm$ 1.6           & \textbf{11.7 $\pm$ 2.1} \\
                          & Genetic & 8.9 $\pm$ 4.3           & 7.2 $\pm$ 4.1  & 6.1 $\pm$ 0.8  & 4.4 $\pm$ 2.1           & 9.4 $\pm$ 1.6  & 5.0 $\pm$ 1.3           & 9.4 $\pm$ 1.6           & \textbf{9.9 $\pm$ 2.1}  \\ \cline{2-10} 
\multirow{4}{*}{IMDB-B}   & Clean   & 52.9 $\pm$ 4.2          & 62.7 $\pm$ 3.9 & 56.7 $\pm$ 4.1 & \textbf{63.7$\pm$ 4.1}  & 52.0 $\pm$ 5.3 & 53.3 $\pm$ 2.4          & 56.7 $\pm$ 4.2          & 62.0 $\pm$ 3.2          \\
                          & PGD     & 52.0 $\pm$ 2.8          & 61.3 $\pm$ 4.4 & 32.0 $\pm$ 5.8 & 61.3 $\pm$ 3.2          & 47.0 $\pm$ 4.5 & 49.6 $\pm$ 2.9          & 56.7 $\pm$ 3.2          & \textbf{61.7 $\pm$ 2.5} \\
                          & Random  & 52.3 $\pm$ 3.2          & 62.3 $\pm$ 3.2 & 45.9 $\pm$ 3.3 & \textbf{62.9 $\pm$ 4.3} & 49.7 $\pm$ 4.0 & 51.3 $\pm$ 1.2          & 55.7 $\pm$ 3.3          & 61.3 $\pm$ 2.9          \\
                          & Genetic & 52.0 $\pm$ 2.8          & 61.0 $\pm$ 2.3 & 52.0 $\pm$ 4.1 & 61.7 $\pm$ 3.8          & 50.9 $\pm$ 4.3 & 52.3 $\pm$ 1.7          & 57.8 $\pm$ 3.3          & \textbf{62.7 $\pm$ 3.2} \\ \cline{2-10} 
\multirow{4}{*}{REDDIT-B} & Clean   & 77.5 $\pm$ 0.8          & 70.0 $\pm$ 1.2 & 72.0 $\pm$ 0.9 & 70.5 $\pm$ 1.6          & 78.5 $\pm$ 1.4 & 77.5 $\pm$ 1.5          & \textbf{80.0 $\pm$ 0.9} & 75.5 $\pm$ 0.6          \\
                          & PGD     & 69.0 $\pm$ 1.4          & 57.5 $\pm$ 2.5 & 62.0 $\pm$ 2.5 & 55.5 $\pm$ 1.9          & 55.5 $\pm$ 1.9 & 57.0 $\pm$ 1.9          & 56.5 $\pm$ 2.5          & \textbf{74.0 $\pm$ 0.9} \\
                          & Random  & 76.0 $\pm$ 0.8          & 69.5 $\pm$ 0.9 & 67.9 $\pm$ 1.2 & 69.0 $\pm$ 1.6          & 76.5 $\pm$ 1.5 & 75.6 $\pm$ 1.4          & \textbf{78.0 $\pm$ 0.6} & 75.0 $\pm$ 0.3          \\
                          & Genetic & 74.0 $\pm$ 0.9          & 69.0 $\pm$ 1.5 & 66.9 $\pm$ 1.4 & 68.5 $\pm$ 1.9          & 55.5 $\pm$ 1.5 & 57.0 $\pm$ 2.5          & 56.5 $\pm$ 1.9          & \textbf{74.5 $\pm$ 0.7} \\ \cline{2-10} 
\multirow{4}{*}{ER\_MD}   & Clean   & 64.0 $\pm$ 3.9          & 61.1 $\pm$ 6.7 & 61.4 $\pm$ 6.1 & 61.1 $\pm$ 5.8          & 66.3 $\pm$ 4.2 & \textbf{69.7 $\pm$ 1.8} & 67.1 $\pm$ 5.4          & 65.9 $\pm$ 5.9          \\
                          & PGD     & 40.4 $\pm$ 3.2          & 44.5 $\pm$ 3.8 & 45.8 $\pm$ 3.3 & 45.9 $\pm$ 2.8          & 37.4 $\pm$ 5.3 & 35.2 $\pm$ 4.5          & 34.1 $\pm$ 6.2          & \textbf{47.9 $\pm$ 3.4} \\
                          & Random  & 55.1 $\pm$ 2.3          & 51.6 $\pm$ 2.8 & 52.4 $\pm$ 3.8 & 49.8 $\pm$ 2.3          & 50.1 $\pm$ 3.8 & 55.1 $\pm$ 2.3          & 52.4 $\pm$ 3.8          & \textbf{58.1 $\pm$ 5.9} \\
                          & Genetic & 60.2 $\pm$ 1.1          & 58.9 $\pm$ 1.8 & 60.2 $\pm$ 0.8 & 59.9 $\pm$ 2.3          & 51.4 $\pm$ 1.8 & 52.4 $\pm$ 4.2          & 55.1 $\pm$ 2.3          & \textbf{61.7 $\pm$ 1.3} \\ \cline{2-10} 
\multirow{4}{*}{MSRC\_9}  & Clean   & 88.6 $\pm$ 3.2          & 88.6 $\pm$ 1.8 & 88.6 $\pm$ 1.8 & \textbf{90.1 $\pm$ 1.1} & 87.1 $\pm$ 2.8 & 89.4 $\pm$ 2.1          & 89.4 $\pm$ 2.8          & 89.7 $\pm$ 1.1          \\
                          & PGD     & 82.6 $\pm$ 2.8          & 84.1 $\pm$ 2.1 & 78.0 $\pm$ 2.9 & 82.6 $\pm$ 2.3          & 54.6 $\pm$ 1.9 & 84.1 $\pm$ 2.3          & 75.8 $\pm$ 3.4          & \textbf{85.6 $\pm$ 3.1} \\
                          & Random  & 85.6 $\pm$ 2.6          & 87.8 $\pm$ 1.3 & 86.3 $\pm$ 1.3 & 87.1 $\pm$ 1.6          & 84.1 $\pm$ 3.2 & 86.8 $\pm$ 1.3          & 86.4 $\pm$ 2.1          & \textbf{89.1 $\pm$ 0.4} \\
                          & Genetic & 85.6 $\pm$ 2.1          & 86.3 $\pm$ 2.8 & 87.1 $\pm$ 1.9 & 87.1 $\pm$ 2.1          & 84.1 $\pm$ 3.2 & 86.3 $\pm$ 1.8          & 86.4 $\pm$ 3.2          & \textbf{89.4 $\pm$ 1.1} \\ \hline
\end{tabular}%
}
\end{table}

%% file: Tables/4_results_features.tex
\begin{table}[]
\caption{Clean and Attacked classification accuracy ($\pm$ standard deviation) of the RS-Pool and other pooling strategies on different graph classification dataset when subject to feautre-based adversarial attacks.}
\label{tab:results_features}
\renewcommand{\arraystretch}{1.1}
\resizebox{\columnwidth}{!}{%
\begin{tabular}{llcccccccc}
\hline
Dataset                   & Attack & Sum                 & Average    & Max        & SAG                 & TopK-P     & PAN-P               & Sort-P     & RS-Pool             \\ \hline
\multirow{3}{*}{PROTEINS} & Clean  & 74.2 ± 3.1          & 70.8 ± 2.2 & 73.2 ± 2.3 & 72.3 ± 3.3          & 73.2 ± 2.9 & \textbf{74.7 ± 1.7} & 72.3 ± 2.3 & 73.5 ± 2.9          \\
                          & PGD    & 68.4 ± 4.0          & 64.6 ± 2.2 & 66.4 ± 4.1 & 64.3 ± 3.6          & 64.9 ± 6.9 & 68.8 ± 2.9          & 66.7 ± 2.9 & \textbf{69.9 ± 4.1} \\
                          & Random & 69.5 ± 5.2          & 67.6 ± 2.9 & 68.4 ± 3.2 & 69.0 ± 4.0          & 68.8 ± 6.5 & 69.2 ± 3.2          & 70.2 ± 2.6 & \textbf{70.9 ± 2.8} \\ \hline
\multirow{3}{*}{D\&D}     & Clean  & 75.1 ± 0.6          & 70.1 ± 0.5 & 74.1 ± 0.6 & 71.3 ± 0.4          & 74.4 ± 1.1 & \textbf{75.7 ± 2.8} & 74.9 ± 1.7 & 74.6 ± 0.7          \\
                          & PGD    & 36.9 ± 3.6          & 36.9 ± 6.8 & 46.4 ± 6.4 & 49.9 ± 1.8          & 28.7 ± 3.7 & 37.5 ± 4.1          & 47.9 ± 4.8 & \textbf{51.8 ± 1.1} \\
                          & Random & 49.0 ± 5.2          & 44.0 ± 6.5 & 45.7 ± 4.6 & 54.1 ± 3.0          & 40.9 ± 5.1 & 51.0 ± 4.6          & 49.7 ± 4.9 & \textbf{54.9 ± 2.3} \\ \hline
\multirow{3}{*}{NCI1}     & Clean  & \textbf{70.6 ± 0.8} & 67.9 ± 1.6 & 68.2 ± 1.2 & 70.4 ± 0.9          & 70.9 ± 0.9 & 70.4 ± 1.7          & 69.5 ± 0.4 & 70.1 ± 1.2          \\
                          & PGD    & 38.6 ± 2.8          & 36.4 ± 3.5 & 37.9 ± 5.3 & 36.6 ± 2.2          & 35.1 ± 2.2 & 38.4 ± 1.8          & 36.2 ± 3.6 & \textbf{39.4 ± 1.9} \\
                          & Random & 41.1 ± 2.0          & 38.1 ± 2.9 & 40.9 ± 5.5 & 40.1 ± 2.9          & 38.5 ± 1.0 & 40.9 ± 2.4          & 38.5 ± 4.8 & \textbf{41.7 ± 1.3} \\ \hline
\multirow{3}{*}{ER\_MD}   & Clean  & 64.0 ± 3.9          & 61.1 ± 6.7 & 61.4 ± 6.1 & 61.1 ± 5.8          & 66.3 ± 4.2 & \textbf{69.7 ± 1.8} & 67.1 ± 5.4 & 65.9 ± 5.9          \\
                          & PGD    & 41.9 ± 3.2          & 45.3 ± 1.4 & 45.3 ± 4.6 & 45.7 ± 3.5          & 45.3 ± 3.8 & 45.3 ± 2.1          & 43.1 ± 3.5 & \textbf{55.1 ± 4.9} \\
                          & Random & 54.3 ± 2.3          & 50.2 ± 4.3 & 52.1 ± 6.4 & 53.9 ± 5.5          & 52.1 ± 4.1 & 51.3 ± 5.2          & 53.6 ± 7.7 & \textbf{56.2 ± 5.1} \\ \hline
\multirow{3}{*}{MSRC\_9}  & Clean  & 88.6 ± 3.2          & 88.6 ± 1.8 & 88.6 ± 1.8 & \textbf{90.1 ± 1.1} & 87.1 ± 2.8 & 89.4 ± 2.1          & 89.4 ± 2.8 & 89.7 ± 1.1          \\
                          & PGD    & 84.8 ± 2.8          & 85.6 ± 4.0 & 85.6 ± 2.8 & 87.1 ± 1.1          & 83.3 ± 1.1 & 85.6 ± 3.9          & 84.1 ± 1.9 & \textbf{88.3 ± 1.1} \\
                          & Random & 86.4 ± 3.7          & 88.6 ± 1.9 & 87.9 ± 2.8 & 89.4 ± 1.1          & 87.1 ± 2.1 & 87.9 ± 3.9          & 88.6 ± 3.2 & \textbf{89.3 ± 1.1} \\ \hline
\end{tabular}%
}
\end{table}

%% file: main_paper/7_Conclusion.tex
In this work, we provided a theoretical analysis of how pooling operations affect the adversarial robustness of GNNs in the context of graph classification. Our findings indicate that different pooling strategies exhibit varying levels of sensitivity to attack types -- some being more robust to targeted perturbations, while others are more vulnerable under global attacks. To address this, we introduced \textit{RS-Pool}, a novel pooling method that uses a scaled version of the dominant singular vector of the node embedding matrix as the graph-level representation. We theoretically demonstrated that RS-Pool can attenuate the impact of adversarial perturbations at the pooling stage, thereby improving the model’s overall robustness. Empirical evaluations on real-world datasets, using different GNN backbones and adversarial attacks, confirm that RS-Pool achieves stronger robustness while maintaining competitive performance on clean data.

\textbf{Limitations.} Our method relies on the existence of a spectral gap in feature matrices, yet gap size can depend on the network depth and graph topology and varies across settings, which warrants further study. Moreover, power iteration pooling scales linearly with edges but may incur nontrivial overhead on very dense graphs, limiting its use in resource constrained applications.

\section*{Acknowledgements}
The computation (on GPUs) was partially enabled by resources provided by the National Academic Infrastructure for Supercomputing in Sweden (NAISS) at Alvis partially funded by the Swedish Research Council through grant agreement no. ``2024/22-309''. J.L. would additionally like to aknowlegde support by the French National Research Agency (ANR) via the ``GraspGNNs'' JCJC grant (ANR-24-CE23-3888). We furthermore want to acknowledge all the reviewers for their feedback that improved our manuscript.

%% file: Appendix/1_theo_gcn.tex
\textbf{Theorem~\ref{theo:main_result}}~Let $f \colon (\mathcal{A}, \mathcal{X}) \rightarrow \mathcal{Y}$ denote a graph-based function composed of $L$ GCN layers, where the weight matrix of the $\ell$-th layer is denoted by $W^{(\ell)}.$ Under a feature-based adversarial attack with perturbation budget $\epsilon$, the robustness of $f$ in the sense of Definition~\ref{def:robustness} satisfies the following:
\begin{itemize}
    \item If $f$ is based on Max pooling, then it is $(\epsilon, \gamma)$ robust with:
    \begin{center}
        $\gamma = \sqrt{\min\{n,d_L\}} \left (\prod_{\ell=1}^L \lVert W^{(\ell)}\lVert \right )  \max_{u \in V}  \hat{w}_u \epsilon.$  
    \end{center}

    \item If $f$ is based on Sum pooling, then it is $(\epsilon, \gamma)$ robust with:
    \begin{center}
        $\gamma = \left (\prod_{\ell=1}^L \lVert W^{(\ell)}\lVert \right ) \sum_{u \in V}  \hat{w}_u \epsilon.$  
    \end{center}

    \item If $f$ is based on Average pooling, then it is $(\epsilon, \gamma)$ robust with:
    \begin{center}
        $\gamma = \frac{\epsilon}{n} \left (\prod_{\ell=1}^L \lVert W^{(\ell)}\lVert \right ) \sum_{u \in V}  \hat{w}_u.$  
    \end{center}
    
\end{itemize}
Here, $\hat{w}_u$ denotes the sum of normalized walks of length $L{-}1$ originating from node $u$, and $d_L$ is the output dimensionality of the final layer.

\begin{proof}

In this proof, we consider that $f$ is a graph-function that is based on $L$ layers of GCN. We recall that the GCN message-passing propagation is formulated for a node $u$ as
\begin{equation}
    h_u^{(\ell)} = \sigma^{(\ell)} (\underset{v \in \mathcal{N}(u) \bigcup \{ u \}}{\Sigma} \frac{W^{(\ell)} h_v^{(\ell-1)}}{ \sqrt{(1 + d_u)(1 + d_v)} }),
\end{equation}
where $W^{(\ell)} \in \mathbb{R}^{d_{\ell-1} \times d_{\ell}}$ is the learnable weight matrix with $d_{\ell}$ being the embedding dimension of layer $\ell$ and $\sigma^{(\ell)}$ is the activation function of $\ell$-th layer. We recall that $h^{(0)} = X \in \mathbb{R}^{n\times d}$ is set to the initial node features.

Similar to \citet{abbahaddou2024bounding}, we denote $X$ as the original node features and denote by $X'$ the perturbed adversarial features. We consider a node $u \in V$, we denote by $h_u$ its representation in the clean graph and $h'_{u}$ its representation in the attacked graph. We consider that the activation functions $(\sigma^{(\ell)})_{1 \leq \ell \leq L}$ are \textit{nonexpansive} (1-Lipschitz continuous). From the work, we have the following result

\begin{align*}
    \lVert h_u^{(L)} - {h'}_{u'}^{(L)} \lVert
   & \leq \prod_{\ell=1}^L \lVert W^{(\ell)}\lVert_{2} \left\lVert \underset{v \in \mathcal{N}(u) \bigcup \{ u \}}{\Sigma} \underset{j \in \mathcal{N}(v) \bigcup \{ v \}}{\Sigma} \ldots \right.\\
   & \hspace{3em} \left.\underset{z \in \mathcal{N}(y) \bigcup \{ y \}}{\Sigma} \frac{X_u - X'_{u}}{\sqrt{(1 + d_u)} (1 + d_w) (1 + d_j) \ldots  (1 + d_y) \sqrt{(1 + d_z)}}   \right\rVert \\
   & \leq \prod_{\ell=1}^L \lVert W^{(\ell)}\lVert  \hat{w}_u \epsilon,
\end{align*}

with $\hat{w}_u$ being the sum of normalized walks of length $(L-1)$ starting from node $u$. 

The previous results give us an idea about the behavior of each node's representation when attacked. In the case of graph classification, an additional pooling operation is added, Specifically:

\begin{equation*}
    h_{\text{graph}}^{(L)} = \text{Pool}\left( \{ h_u^{(L)} \}_{u \in V} \right).
\end{equation*}

Hence this proof's goal is to analyze the following quantity
\begin{equation*}
    \lVert h_{\text{graph}}^{(L)} - {h'}_{\text{graph}}^{(L)} \rVert.
\end{equation*}

Let's consider the \textbf{Sum pooling} operation, that can be written as
\begin{equation*}
    h_{\text{graph}}^{(L)} = \sum_{u \in V} h_u^{(L)}.
\end{equation*}

We have the following
\begin{align*}
    \lVert h_{\text{graph}}^{(L)} - {h'}_{\text{graph}}^{(L)} \rVert 
    &= \left\lVert \sum_{u \in V} \left( h_u^{(L)} - {h'}_u^{(L)} \right) \right\rVert \\
    &\leq \sum_{u \in V} \lVert h_u^{(L)} - {h'}_u^{(L)} \rVert \quad \text{(by triangle inequality)} \\
    &\leq \sum_{u \in V} \left( \prod_{\ell=1}^L \lVert W^{(\ell)} \rVert \right) \lVert \hat{w}_u \rVert \epsilon \\
    &= \left( \prod_{\ell=1}^L \lVert W^{(\ell)} \rVert \right) \epsilon \sum_{u \in V}  \hat{w}_u.
\end{align*}

For the case of the \textbf{Average pooling} operation, that can be written as
\begin{equation*}
    h_{\text{graph}}^{(L)} = \frac{1}{|V|} \sum_{u \in V} h_u^{(L)}.
\end{equation*}

We have the following analysis

\begin{align*}
    \lVert h_{\text{graph}}^{(L)} - {h'}_{\text{graph}}^{(L)} \rVert 
    &= \left\lVert \frac{1}{|V|} \sum_{u \in V} \left( h_u^{(L)} - {h'}_u^{(L)} \right) \right\rVert \\
    &\leq \frac{1}{|V|} \sum_{u \in V} \lVert h_u^{(L)} - {h'}_u^{(L)} \rVert \quad \text{(by triangle inequality)} \\
    &\leq \frac{1}{|V|} \left( \prod_{\ell=1}^L \lVert W^{(\ell)} \rVert \right) \epsilon \sum_{u \in V} \hat{w}_u.
\end{align*}

In the case of \textbf{Max pooling}: Let's denote $\Delta = H^{(L)} - {H'}^{(L)}$. Since the magnitude of a single entry of a row can never exceed the Euclidean norm of the whole row, which can be formulated as for any column $j$ of our difference matrix that $\mid \Delta_{u,j} \mid \leq \lVert \Delta_{u, :} \lVert$ For Max poling, we have the following

\begin{align*}
    \lVert h_{\text{graph}}^{(L)} - {h'}_{\text{graph}}^{(L)} 
    \rVert_2 & \le \sqrt{d_L}\max_u\lVert \Delta_{u:}\rVert_2 \\
    & \le \epsilon \sqrt{d_L}\left( \prod_{\ell=1}^L \lVert W^{(\ell)} \rVert \right)\max_{u \in V}  \hat{w}_u.
\end{align*}

We additionally note that we have the following
\begin{align}\label{eq:max_inequality}
    &\lVert h_{\text{graph}}^{(L)} - {h'}_{\text{graph}}^{(L)} 
    \rVert^2 = \sum_{j=1}^{d_L}\max_{u}\mid  \Delta_{uj}\mid^2 
    \le \sum_{u,j}\mid\Delta_{uj}\mid^2 = \lVert \Delta\rVert_F^2.
\end{align}
And we have that
\begin{align*}
  \lVert \Delta\rVert_F \leq \epsilon \left( \prod_{\ell=1}^L \lVert W^{(\ell)} \rVert \right)\left(\sum_u (\hat{w}^{(L)}_u)^2\right)^{1/2}.
\end{align*}

From the two derived inequalities, we conclude that in the case of Max pooling we have
\begin{align*}
    \lVert h_{\text{graph}}^{(L)} - {h'}_{\text{graph}}^{(L)} \rVert \leq \epsilon \sqrt{\min\{n,d_L\}}\left( \prod_{\ell=1}^L \lVert W^{(\ell)} \rVert \right)\max_{u \in V}  \hat{w}_u.
\end{align*}

By taking into account the expectancy (as shown in Definition \ref{equation:robustness_definition}), we get the desired results.

\end{proof}

%% file: Appendix/2_theo_gin.tex
\textbf{Theorem~\ref{theo:result_GIN}}~Let $f\colon (\mathcal{A}, \mathcal{X}) \rightarrow \mathcal{Y}$ be composed of $L$ GIN-layers (with its internal parameter $\zeta = 0$) and let $W^{(\ell)}$ denote the weight matrix of the $\ell$-th MLP layer. We consider the input node feature space to be bounded, \ie $\lVert X \rVert_2 < B$ for some $B\in \mathbb{R}.$ Under a feature-based adversarial attack with perturbation budget $\epsilon$, the model $f$ is $(\epsilon, \gamma)$-robust with respect to Definition~\ref{def:robustness}, with:
\begin{itemize}[leftmargin=4em]
    \item If $f$ is based on Max pooling: $\gamma = \prod_{\ell=1}^L \lVert W^{(\ell)} \rVert \left( B  L \left(\max_{u \in V} \deg(u)\right) + \epsilon \right).$

    \item If $f$ is based on Sum pooling: $\gamma = \prod_{\ell=1}^L \lVert W^{(\ell)} \lVert \left( 2B  L |E| + n \epsilon \right).$

    \item If $f$ is based on Average pooling: $\gamma = \sqrt{n, d_{L}} \big(\prod_{\ell=1}^L \lVert W^{(\ell)} \lVert \big) \left( \frac{2B  L  |E|}{n} + \epsilon \right).$

\end{itemize}
Here, $|E|$ is the number of edges, $n$ number of nodes and $d_L$ the dimensionality of the final layer $L$.

\begin{proof}
In this proof, we consider that $f$ is graph-based function that is based on $L$ GIN-layers (with a parameter $\zeta = 0,$ usually denoted as $\epsilon$, but changed to avoid confusion with our attack budget). The GIN message-passing propagation process can be written for a node $u$ as:
\begin{equation*}
    h_u^{(\ell + 1)} = T^{(\ell + 1)}( (1 + \zeta) h_u^{(\ell)} +  \underset{v \in \mathcal{N}(u)}{\Sigma} h_v^{(\ell)}),
\end{equation*}

with $T$ denoting a Neural Networks (usually a MLP) and $\zeta$ denotes the parameter of the GIN.  We recall that $h^{(0)} = X \in \mathbb{R}^{n\times d}$ is set to the initial node features.

Similar to the previous proof, we base our analysis on previous work \citep{abbahaddou2024bounding}, we denote $X$ as the original node features and denote by $X'$ the perturbed adversarial features. We consider a node $u \in V$, we denote by $h_u$ its representation in the clean graph and $h'_{u}$ its representation in the attacked graph. We recall that in our problem setup, we consider that the activation functions $(\sigma^{(\ell)})_{1 \leq \ell \leq L}$ are \textit{nonexpensive} (1-Lipschitz continuous). 

We use the same assumptions as the one considered in \citep{abbahaddou2024bounding}. Specifically, we consider that the input feature space $\mathcal{H}_0$ is bounded, thus each hidden space $\mathcal{H}_i$ of the iterative process of message passing is bounded and let $B = \underset{\ell \leq L}{\max} B_{\ell}$ be its global maximum bound. We additionally consider that GIN-parameter $\zeta \approx 0$ (which is very frequent in the literature). For a node $u \in V$, we have the following result:
\begin{equation*}
    \lVert h_u^{(\ell + 1)} - {h'}_{u'}^{(\ell + 1)} \lVert \leq \prod_{\ell=1}^L \lVert W^{(\ell)}\lVert \left(B  L  \deg(u) + \epsilon \right).
\end{equation*}

From this perspective, let's consider the case of graph classification. We omit the definition of the different pooling as this was stated in the previous proof. We start by considering the \textbf{Sum pooling} operation:
\begin{align*}
\lVert h_{\text{graph}}^{(L)} - {h'}_{\text{graph}}^{(L)} \lVert &= \lVert \sum_{u \in V} \left( h_u^{(L)} - {h'}_u^{(L)} \right) \lVert \\
&\leq \sum_{u \in V} \lVert h_u^{(L)} - {h'}_u^{(L)} \lVert \quad \text{(by the triangle inequality)} \\
&\leq \prod_{\ell=1}^L \lVert W^{(\ell)} \lVert \sum_{u \in V} \left( B  L  \deg(u) + \epsilon \right) \\
&= \prod_{\ell=1}^L \lVert W^{(\ell)} \lVert \left( B  L  \sum_{u \in V} \deg(u) + | V| \epsilon \right).
\end{align*}

In the case of undirected graph, since $\sum_{u \in V} \deg(u) = 2 | E|$, based on the previous result, we can write:
\begin{equation*}
    \lVert h_{\text{graph}}^{(L)} - {h'}_{\text{graph}}^{(L)} \lVert \leq \prod_{\ell=1}^L \lVert W^{(\ell)} \lVert \left( 2B  L  | E | + | V | \epsilon \right).
\end{equation*}

Similar for the case of \textbf{Average pooling}, we have:
\begin{align*}
\lVert h_{\text{graph}}^{(L)} - {h'}_{\text{graph}}^{(L)} \lVert &= \frac{1}{|V|} \lVert \sum_{u \in V} \left( h_u^{(L)} - {h'}_u^{(L)} \right) \lVert \\
&\leq \frac{1}{|V|} \sum_{u \in V} \lVert h_u^{(L)} - {h'}_u^{(L)} \lVert \\
&\leq \prod_{\ell=1}^L \lVert W^{(\ell)} \lVert \left( \frac{2B  L  | E|}{| V|} + \epsilon \right).
\end{align*}

In the case of \textbf{Max pooling} operation, following the derivated Inequality \ref{eq:max_inequality}, we can directly get:

\begin{equation*}
    \lVert h_{\text{graph}}^{(L)} - {h'}_{\text{graph}}^{(L)} \lVert \leq \sqrt{n, d_{L}} \big( \prod_{\ell=1}^L \lVert W^{(\ell)} \lVert \big) \left( B  L \left( \max_{u \in V} \deg(u)\right) + \epsilon \right).
\end{equation*}

By taking into account the expectancy (as shown in Definition \ref{equation:robustness_definition}), we get the desired results.

\end{proof}

%% file: Appendix/3_theo_r_pool.tex
\textbf{Theorem \ref{theo:robustness_of_rpool}} Let $f \colon (\mathcal{G}, \mathcal{X}) \rightarrow \mathcal{Y}$ denote a graph-based function composed of $L$ GCN layers and using our RS-Pool, where the weight matrix of the $i$-th layer is denoted by $W^{(i)}.$ Under a feature-based adversarial attack with perturbation budget $\epsilon$, $f$ is $(\epsilon, \gamma)$-robust in the sense of Definition~\ref{def:robustness} with:
$$
    \gamma = \frac{\tau \sqrt2 \epsilon}{\sigma_1 - \sigma_2} \Bigl(\prod_{\ell=1}^L \lVert W^{(\ell)}\lVert\Bigr) 
   \sum_{u=1}^n (\hat{w}_u)^2,
$$
with $\hat{w_u}$ denoting the sum of normalized walks of length $(L-1)$ starting from node $u,$ and $\sigma_1 \neq \sigma_2$ being the two dominant singular values. 

\begin{proof}
In this proof, we consider that $f$ is a graph-function that is based on $L$ layers of GCN. Similar to previous proof, we denote by $X'$ the perturbed node features, and $h'_u{}^{(\ell)}$ the corresponding node embedding of $u$ at layer~$\ell$.  Following~\cite{abbahaddou2024bounding}, for each node $u$ we have:
\begin{equation}
\label{eq:node-level-bound}
    \lVert h_u^{(L)} - h'_u{}^{(L)}\lVert \leq
    \Bigl(\prod_{\ell=1}^L \lVert W^{(\ell)}\lVert\Bigr) \hat{w}_u \epsilon,
\end{equation}

Thus, to relate $\lVert H^{(L)} - H'^{(L)}\lVert$ in operator norm
\begin{align*}
    \lVert H^{(L)} - H'^{(L)} \lVert_F^2 = \sum_{u=1}^n \lVert h_u^{(L)} - h'_u{}^{(L)}\lVert^2 \leq \Bigl(\prod_{\ell=1}^L \lVert W^{(\ell)}\lVert\Bigr)^2\,\epsilon^2
   \sum_{u=1}^n (\hat{w}_u)^2.
\end{align*}
Consequently, we have
\begin{align}\label{eq:matrix-norm-bound}
    \lVert H^{(L)} - H'^{(L)} \lVert \leq \lVert H^{(L)} - H'^{(L)} \lVert_F \leq \Bigl(\prod_{\ell=1}^L \lVert W^{(\ell)}\lVert\Bigr) \epsilon
   \sum_{u=1}^n (\hat{w}_u)^2.
\end{align}

We consider now our proposed pooling $\textrm{RS-Pool}$, which consists of considering the dominant right singular vector of $H^{(L)}$. In this perspective, let:
\begin{align*}
    S = \operatorname{span}\{v_{1}(H^{(L)}\} \text{ and } S' = \operatorname{span}\{v_{1}(H'^{(L)}\} \text{ and accordingly: } \theta = \Theta(S,S')\in\bigl[0,\tfrac{\pi}{2}\bigr], 
\end{align*}

with $\theta$ being the principle angle between the two vectors. Note that we ensure to always take the smaller of the two possible angles between the two dominant vectors (by controlling the signs). Building on Wedin's sin-$\Theta$ theorem~\cite{wedin1972perturbation}, we can write:

\begin{align*}
    \sin\theta \leq \frac{\lVert H - H' \lVert}{\sigma_1(H) - \sigma_2(H)}.
\end{align*}

From another perspective, for our considered subspaces $S$ and $S'$, when considering the principal angle, we have the following:
\begin{align*}
\cos\theta = \max_{\substack{x\in S, y\in S'\\
                \lVert x \rVert =\lVert y \rVert =1}}
                 \langle x,y \rangle
           = \lvert \langle v_{1},v'_{1}\rangle \rvert.
\end{align*}

We may assume $\langle v_1, v_1' \rangle \ge 0$, since the sign of the dominant singular vector can be chosen by definition. From a different perspective, we can express as follows:

\begin{align*}
    \lVert v_{1} - v_{1}' \rVert^{2} = \langle v_{1}-v'_{1},v_{1} - v'_{1}\rangle & = 2 - 2 \langle v_1, v'_{1} \rangle \\
    & = 2 - 2 \cos \theta \\
    & = 4 \sin^2(\theta / 2).
\end{align*}

We additionally have the following (note that $ \theta \in [0,\tfrac{\pi}{2}])$:
\begin{align*}
    \sin \theta = 2 \sin \frac{\theta}{2} \cos \frac{\theta}{2} \Rightarrow \sin \frac{\theta}{2} = \frac{\sin \theta}{2 \cos \frac{\theta}{2}} \leq \frac{\sin \theta}{2 \cos \frac{\sqrt2}{2}} = \frac{\sin \theta}{\sqrt2}.
\end{align*}

From the two computed elements, we find that:
\begin{align*}
    \lVert v_{1} - v_{1}' \rVert^{2} &\leq 2 \sin(\theta / 2) \\
    & \leq 2 \frac{\sin \theta}{\sqrt2} \\
    & \leq \sqrt2 \sin \theta \\
    & \leq \sqrt2 \frac{\lVert H - H' \lVert}{\sigma_1(H) - \sigma_2(H)} \\
    & \leq \frac{\sqrt2 \epsilon}{\sigma_1 - \sigma_2} \Bigl(\prod_{\ell=1}^L \lVert W^{(\ell)}\lVert\Bigr) 
   \sum_{u=1}^n (\hat{w}_u)^2.
\end{align*}

By also incorporating the constant $\tau \in \mathbb{R}_{>0}$, we obtain the final bound:
\begin{align*}
    \gamma = \frac{\tau \sqrt2 \epsilon}{\sigma_1 - \sigma_2} \Bigl(\prod_{\ell=1}^L \lVert W^{(\ell)}\lVert\Bigr) 
   \sum_{u=1}^n (\hat{w}_u)^2.
\end{align*}

\textbf{On the special case of singular value degeneracy.} In the previous analysis, we consider that the dominant singular value doesn't have a multiplicity higher than $1$. We consider that this is the most dominant case in practice. Specifically, we note that RS-Pool is applied to the final node embedding matrix produced by the message-passing scheme. At this stage, having a dominant singular value with high multiplicity is highly unlikely, even in symmetric graphs, because the learned embeddings do not depend just on the graph structure but also on initial node features and the GNN parameters, which typically break such dominant symmetries. Moreover, GNNs are known for their inherent smoothing tendencies, which lead to rank-reduction in the feature matrix and result in a non-trivial spectral gap in practice.

Nonetheless, extending to the case where the dominant singular value is of multiplicity $r$ shall result in an additional scalar $\sqrt(r)$ (rather than $\sqrt(2)$) in the derived upper-bound $\gamma$ and the considered spectral gap is $\sigma_r - \sigma_{r+1}$.

\end{proof}

\textbf{Corollary \ref{corollary:effect_of_tau}} Let $f: (\mathcal{G}, \mathcal{X}) \rightarrow \mathcal{Y}$ be the graph-based function considered in Theorem~\ref{theo:robustness_of_rpool}. Then $f$ is $(\epsilon, \gamma')$-robust with:
    \begin{center}
        $\gamma' = \min\{\gamma, 2 \tau\}, $
    \end{center}
where $\gamma$ is the bound derived in Theorem~\ref{theo:robustness_of_rpool}.

\begin{proof}
Let's consider the model $f: (\mathcal{G}, \mathcal{X}) \rightarrow \mathcal{Y}$ that uses our proposed $\textrm{RS-Pool}$. 

From Theorem~\ref{theo:robustness_of_rpool}, we have seen that $f$ is $(\epsilon, \gamma)$-robust with: 
\begin{align*}
\gamma = \frac{\tau \sqrt2 \epsilon}{\sigma_1 - \sigma_2} \Bigl(\prod_{\ell=1}^L \lVert W^{(\ell)}\lVert\Bigr) 
   \sum_{u=1}^n (\hat{w}_u)^2.
\end{align*}

Using the triangular inequality, we have the following:
\begin{align*}
    \lVert h_G - h_{G'} \rVert & = \lVert \textrm{RS-Pool}(H) - \textrm{RS-Pool}(H') \rVert \\
    & =  \lVert \tau v_1(H) - \tau v_1(H') \rVert \\
    & = \tau \lVert v_1(H) - v_1(H') \rVert \\
    & \leq \tau \bigl[ \lVert v_1(H) \rVert + \lVert v_1(H') \rVert \bigr]  \\
    & \leq 2 \tau.
\end{align*}

And therefore by combining the two provided bounds, we can write that $f$ is $(\epsilon, \gamma')$-robust with:
\begin{center}
    $\gamma' = \min\{\gamma, 2 \tau\}. $
\end{center}
\end{proof}

%% file: Appendix/4_proof_remark.tex
\textbf{Lemma \ref{lemma:permutation_invariance}.} RS-Pool is permutation invariant; that is, for any pair of isomorphic graphs $G$ and $G_{\Pi}$ with its corresponding permutation matrix $\Pi$, we have $f(A, X) = f(A_{\Pi}, X_{\Pi})$.

\begin{proof}
Let $X \in \mathbb{R}^{n \times d}$ be a matrix, let's consider its SVD decomposition:
\begin{align*}
    X = U \Sigma V^T.
\end{align*}

Let's $X'$ be a permuted version of the adjacency matrix $X$, and let $P \in R^{n \times n}$ be the corresponding permutation matrix. Since by definition $P^TP = PP^T = I_n$, we can write the following:
\begin{align*}
    (PX)^T(PX) = X^TP^TPX = X^TX.
\end{align*}

From the previous equality, we observe that $X^T X$ and $(PX)^T (PX)$ are identical. Consequently, they share the same largest eigenvalue and corresponding eigenspace. Thus, any unit-norm dominant eigenvector of $PX$ must lie in the one-dimensional subspace spanned by $v_1(X)$. By enforcing a consistent sign convention during pooling (e.g., requiring the first non-zero entry to be positive), we obtain:
\begin{align*}
\text{RS-Pool}(X’) = \text{RS-Pool}(PX) = \tau v_1(PX) = \tau v_1(X) = \text{RS-Pool}(X).
\end{align*}

As a result, we obtain permutation invariance in our pooling operation when $X$ represents the output of the message-passing stage.
\end{proof}

%% file: Appendix/5_Additional_results.tex
\subsection{Convergence Analysis}\label{sec:convergence_analysis}
As outlined in Section~\ref{sec:proposed_pooling}, our proposed RS-Pool method relies on estimating the dominant singular vector of an input matrix using the iterative power method. This classical algorithm begins with a randomly initialized vector and performs a series of matrix-vector multiplications, progressively aligning the estimate with the dominant singular vector of the matrix. The number of iterations, denoted by $K$, directly influences the quality of this approximation and, by extension, the effectiveness of the pooling operation.

In this section, we empirically investigate how the choice of $K$ impacts the accuracy of the estimated singular vector. To this end, we compare the vector obtained after $K$ iterations of the power method to the  dominant singular vector computed via the more precise  full SVD implemented in the gesdd routine of the LAPACK package. We use the $\ell_2$-norm to measure the distance between the estimated and true singular vectors. This allows us to quantify the convergence behavior of the power method and assess how quickly a sufficiently accurate approximation can be achieved.

\begin{figure}[H]
\centering
\begin{minipage}{.24\linewidth}
  \centering
  \includegraphics[width=\linewidth]{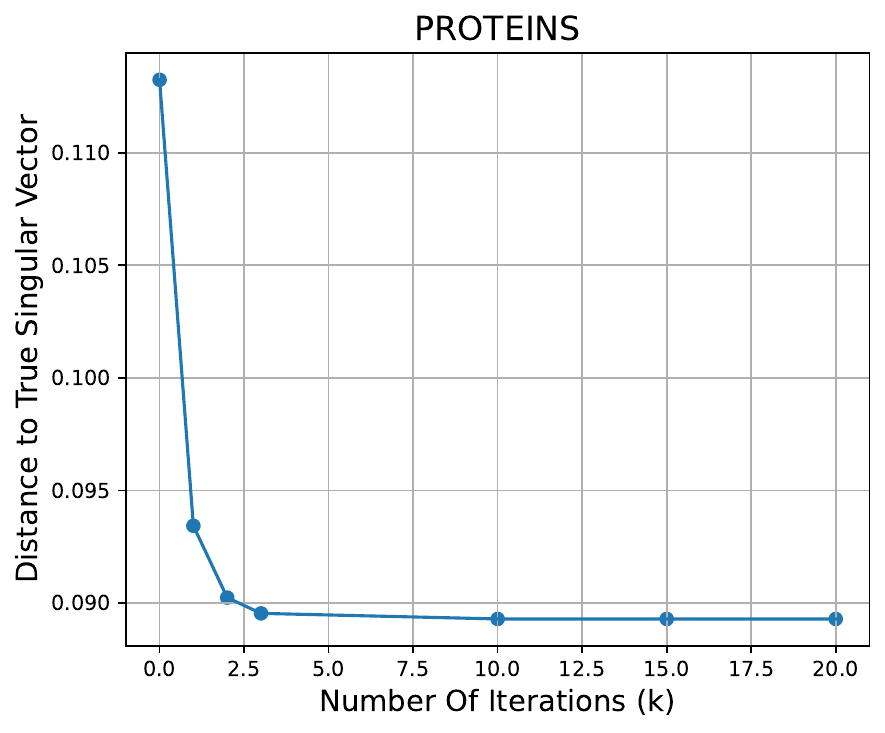}
\end{minipage}
\hfill
\begin{minipage}{.24\linewidth}
  \centering
  \includegraphics[width=\linewidth]{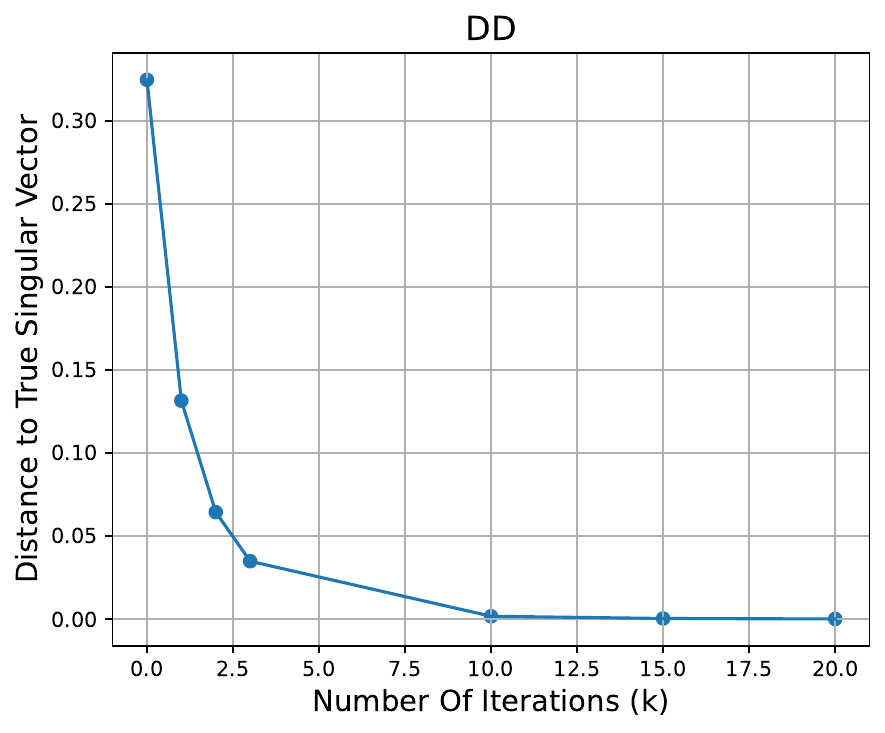}
\end{minipage}
\hfill
\begin{minipage}{.24\linewidth}
  \centering
  \includegraphics[width=\linewidth]{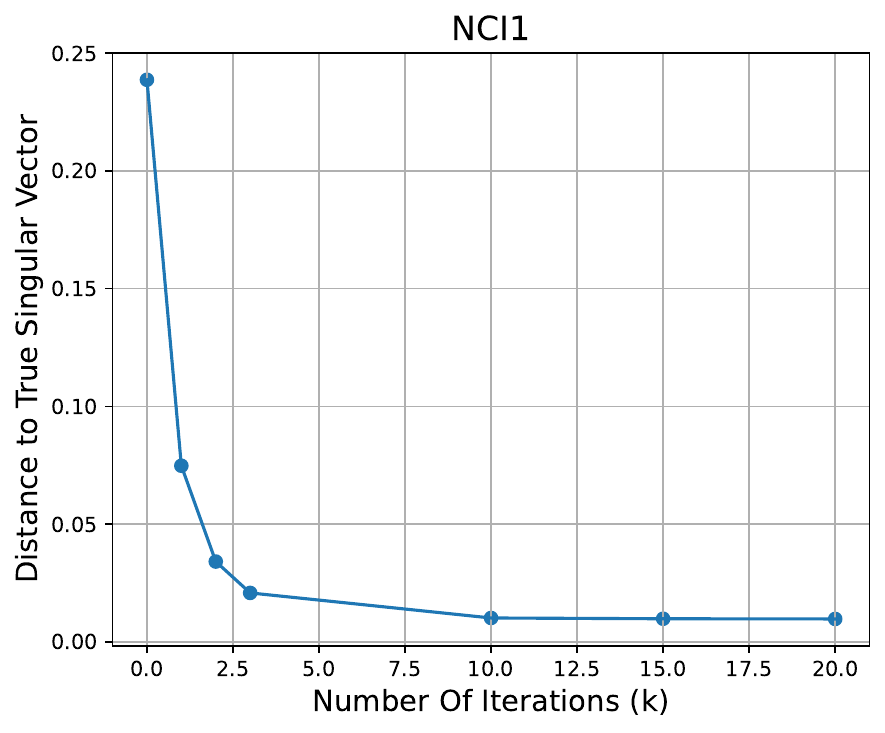}
\end{minipage}
\hfill
\begin{minipage}{.24\linewidth}
  \centering
  \includegraphics[width=\linewidth]{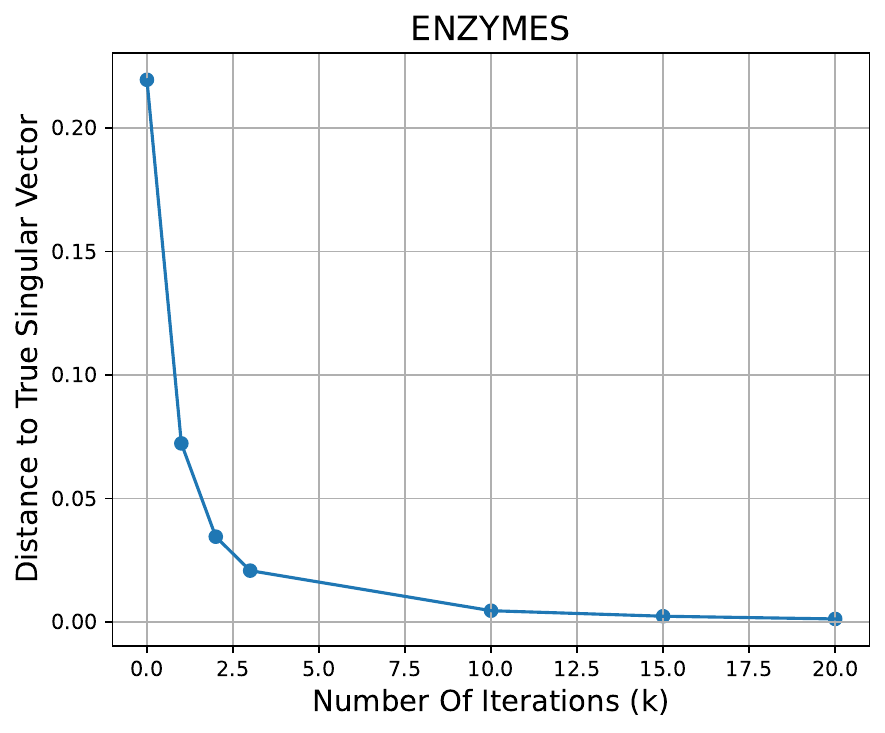}
\end{minipage}
\caption{Effect of the number of iterations on the convergence of the iterative power method. The convergence is computed through the $\ell_2$ distance between the estimated and true dominant singular vectors (obtained via full SVD).}
\label{fig:convergence_analysis}
\end{figure}

Figure~\ref{fig:convergence_analysis} summarizes the convergence behavior of our method across multiple datasets. We observe that in most cases, as few as 2 to 5 iterations are sufficient to obtain a reliable estimate of the dominant singular vector. This rapid convergence ensures that RS-Pool achieves stable performance in terms of both clean and adversarial accuracy, confirming its computational efficiency with a small number of iterations $K$. Moreover, the results indicate that the spectral gap $\sigma_1(X) - \sigma_2(X)$ is sufficiently large, which, according to Theorem~\ref{theo:robustness_of_rpool}, guarantees robustness to perturbations within a budget $\epsilon$.

\subsection{Time Analysis}\label{app:time_analysis}
In this section, we analyze the computational cost of our proposed RS-Pool method, with a particular focus on its time complexity. Building on the convergence analysis presented earlier, we examine how the number of power iterations $K$ affects training time, clean accuracy, and adversarial robustness.

Table~\ref{tab:num_iter_analysis_proteins} reports results on the \textsc{PROTEINS} dataset. We observe that increasing the number of iterations yields only marginal improvements in clean accuracy. Similarly, the attack success rate decreases by less than $2\%$ when increasing $K$ from $1$ to $10$. In contrast, training time grows significantly due to the additional cost of repeated matrix-vector multiplications in the power iteration.

These results underscore the importance of selecting an appropriate number of iterations to balance model performance and computational efficiency. Taken together with the convergence properties discussed earlier, those findings support that a small number of iterations (e.g., $K = 2$ to $5$) is sufficient to achieve strong robustness with minimal overhead.

\begin{table}[h]
\tiny
\caption{Training time and Performance in terms of both clean and attack success rate for different number of iterations for the PROTEINS and NCI1 dataset for $\epsilon=0.2$.}
\label{tab:num_iter_analysis_proteins}
\vskip 0.01in
\begin{center}
\begin{small}
\begin{sc}
\begin{tabular}{llcccc}
\hline
Dataset           &                      & $K = 1$ & $K = 2$  & $K = 5$  & $K = 10$  \\ \hline
\multirow{3}{*}{PROTEINS} & Training Time (in s) & 123.57  & 148.96 & 226.63 & 350.95  \\
                  & Clean Accuracy       & 73.36   & 73.95  & 73.51  & 73.06   \\
                  & Attack Success Rate  & 21.04   & 21.25  & 20.50  & 19.89   \\ \hline
\end{tabular}
\end{sc}
\end{small}
\end{center}
\end{table}

We now focus on comparing the training time of the proposed RS-Pool against the benchmark pooling methods that were considered both in our theoretical study and our experimental results in Section \ref{sec:exp_results}.

\begin{table}[h]
\caption{Mean training time (seconds) of the proposed RS-Pool in comparison to other pooling benchmarks on the different datasets.}
\small
\label{tab:time_analysis}
\begin{center}
\begin{tabular}{lcccccccc}
\hline
Dataset     & Sum     & Avg     & Max     & SAG     & TopK-P  & Pan-P   & Sort-P  & RS-Pool \\ \hline
PROTEINS    & 56.95   & 56.50   & 60.78   & 81.27   & 71.32   & 68.98   & 71.59   & 148.96  \\
D\&D          & 1187.10 & 1176.93 & 1182.77 & 1206.33 & 1206.44 & 1210.92 & 1209.17 & 1293.93 \\
NCI1        & 131.19  & 132.88  & 138.15  & 203.11  & 180.05  & 170.05  & 186.97  & 249.88  \\
ENZYMES     & 20.42   & 20.57   & 21.54   & 30.76   & 27.54   & 26.34   & 28.28   & 46.30   \\
IMDB-B & 29.05   & 29.42   & 30.88   & 46.50   & 41.43   & 39.04   & 43.32   & 75.61   \\
ER\_MD      & 11.40   & 11.52   & 12.13   & 18.26   & 16.21   & 15.15   & 16.88   & 29.23   \\
MSRC\_9     & 6.62    & 6.71    & 7.06    & 9.93    & 8.96    & 8.39    & 9.31    & 18.43   \\ \hline
\end{tabular}
\end{center}
\end{table}

\subsection{Additional Results - GCN}\label{app:additional_results}
Extending the analysis presented in Section~\ref{sec:exp_results}, we continue our empirical evaluation of the upper bound $\gamma$ across the remaining pooling methods. Consistent with earlier findings, RS-Pool exhibits a smaller distance between the pooled representations of clean and perturbed graphs compared to all other methods. This reduced discrepancy correlates with a lower attack success rate, highlighting RS-Pool’s improved adversarial robustness.

\begin{figure}[h]
    \centering
    \includegraphics[width=\linewidth]{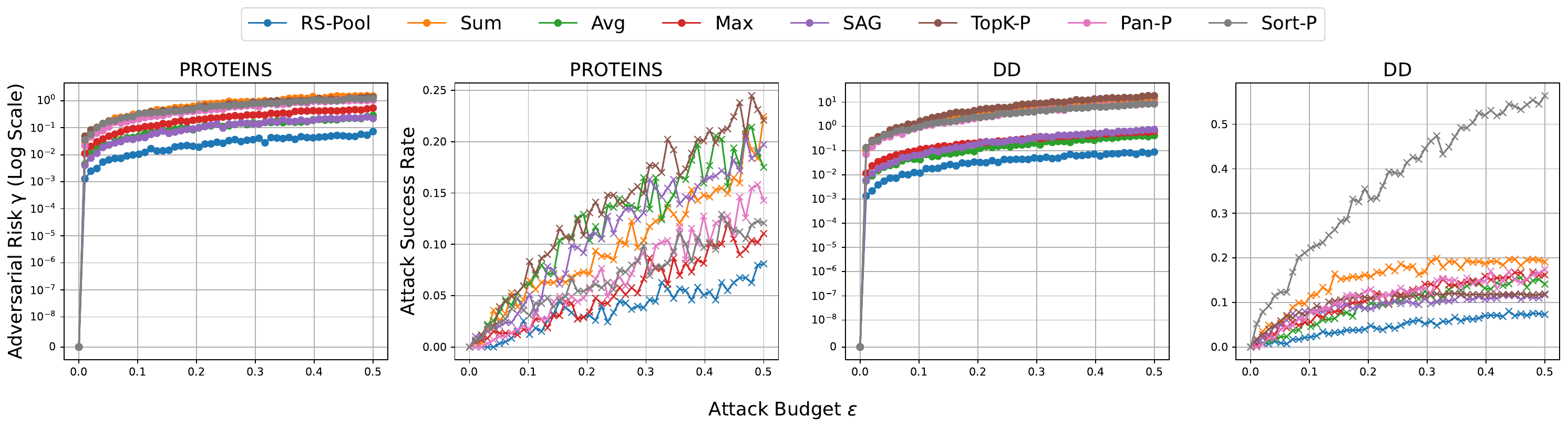}
    \caption{Extension of Figure \ref{fig:analysis_of_risk}: Empirical estimation of adversarial risk $\gamma$ and corresponding attack success rate on PROTEINS ((a), (b)) and D\&D ((c), (d)).}
    \label{fig:risk_quantity_all_poolings}
\end{figure}

\subsection{Additional Results - GIN}\label{app:results_gin}
We further want to evaluate our proposed RS-Pool when considering a different backbone architecture, specifically Graph Isomorphism Networks (GIN). In this setup, we consider a model based on $2$ GIN layers as the message-passing framework. Table~\ref{tab:results_gin} reports the mean and standard deviation of both clean and adversarial accuracy under PGD attacks with a perturbation budget of  $\epsilon = 0.3$. Notably, we observe that in the case of GIN, increasing the number of power iteration steps used to approximate the dominant singular vector improves performance. As a result, we increase the number of iterations from $k = 2$ (used with GCN) to $k = 5$. Overall, RS-Pool consistently outperforms the baseline pooling methods, demonstrating similar robustness improvements to those observed with GCN. These results support our claim that RS-Pool is model-agnostic and can be effectively integrated into different GNN architectures for graph classification.

\input{Tables/2_gin_table}

\subsection{Additional Results - Adversarial Defenses}\label{app:results_defense}

Since RS-Pool is model-agnostic, we investigate its effectiveness when combined with existing adversarial defense strategies. We consider four representative defense categories. First, a \textit{post-processing} defense via adversarial training, where adversarial examples are generated and used to retrain the model to improve robustness. We additionally consider Randomized smoothing, which consists of generating a number of points within the input and then take the majority vote as a classification. We afterwards consider a set of \textit{architecture-level} defense using both R-GCN, which modifies the message-passing mechanism to improve stability under perturbations, and SoftMedian, which introduces a robust aggregation scheme that down-weights neighbors based on their distance to the dimension-wise median, mitigating the impact of outliers or malicious inputs.

Table~\ref{tab:results_defense} provides the mean clean and attacked accuracy for the different considered pooling benchmarks when subject to the PDG adversarial attack. We observe similar insights as in the case of the GCN and GIN, where our proposed RS-Pool out-perform all the other poolings in terms of attacked accuracy showcasing therefore its ability to work with different undelrying backbone GNNs.

\input{Tables/3_defense_methods}

\subsection{Bit-Flip Attacks (BFAs)}\label{app:bit_flip_attacks}
Recently, Bit-Flip attacks have emerged as a new paradigm for attacking and consequently evaluating the robustness of deep learning models at the hardware and memory level. Unlike conventional adversarial attacks that perturb inputs or gradients, bit-flip attacks directly manipulate the binary representation of a model’s parameters in memory. By flipping only a few critical bits, an adversary can drastically alter the network’s behavior, causing severe accuracy degradation or targeted misclassification. While such attacks have been extensively studied in the context of convolutional and feedforward architectures, their implications for Graph Neural Networks (GNNs) are only beginning to be understood. Recent work \cite{kummer2024on, kummer2025on} have shown that even minimal bit-level corruption can disrupt the delicate balance of the message-passing propagation mechanism.

In this perspective, and to further demonstrate the robustness and validity of our proposed RS-Pool, we extend our evaluation to include this line of attack. Specifically, we adapt the \emph{Injectivity Bit-Flip Attack} (IBFA) by removing its quantization step, allowing it to operate directly on our floating-point model. We employ the same dataset and follow an identical experimental setup, using the GIN architecture and setting the attack parameter $k = 5$, consistent with the configuration used by the original authors (on the OGB-molHIV molecular dataset). The results of this analysis, summarized in Table~\ref{tab:bitflip_results}, indicate that indeed RS-Pool consistently outperforms all other pooling baselines, further validating its resilience to bit-level adversarial perturbations.

\input{Tables/5_bit_flip_attacks}

\subsection{On the Tightness of Upper-Bounds}
\begin{figure}[h]
    \centering
    \includegraphics[width=\linewidth]{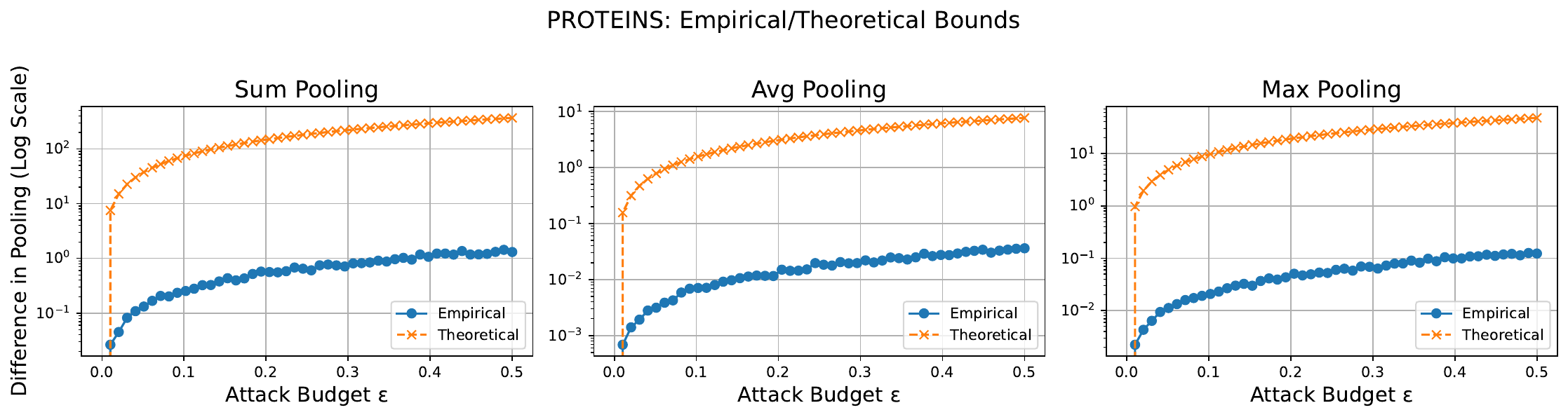}
    \includegraphics[width=\linewidth]{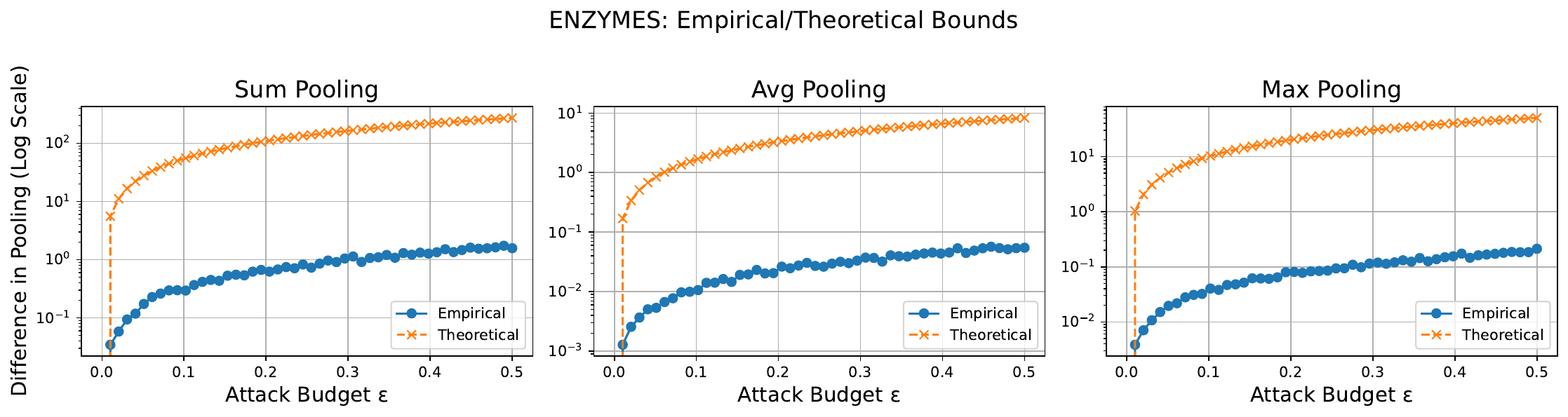}
    \caption{Analysis of the Upper-Bounds provided in Theorem \ref{theo:main_result} and their empirical counter-part for different values of $\epsilon$ on the PROTEINS (Upper) and ENZYMES (Lower) Dataset.}
    \label{fig:theoretical_bound}
\end{figure}

In this section, we assess the tightness of the theoretical upper bound $\gamma$ established in Theorem~\ref{theo:main_result}. To this end, we compare the theoretical bound with empirical estimates computed as the norm difference between the pooled representations of clean and perturbed graphs for different values of $\epsilon$. The results are shown in Figure~\ref{fig:theoretical_bound}. Across all pooling methods, we observe a clear correlation between the empirical measurements and the theoretical bound, providing empirical support for the validity of our theoretical analysis.

%% file: Tables/2_gin_table.tex
\begin{table}[h]
\small

\caption{Clean and Attacked classification accuracy ($\pm$ standard deviation) of the RS-Pool and other pooling strategies applied on the GIN model when subject to the PGD adversarial attacks.}
\label{tab:results_gin}
\renewcommand{\arraystretch}{1.5}
\vspace{1em}
\resizebox{\columnwidth}{!}{%
\begin{tabular}{llcccccccc}
\cline{3-10}
                          &          & Sum                     & Average                 & Max                     & SAG            & TopK-P         & PAN-P          & Sort-P                  & RS-Pool                 \\ \hline
\multirow{2}{*}{PROTEINS} & Clean    & 75.6 $\pm$ 3.1          & 69.3 $\pm$ 1.5          & \textbf{75.8 $\pm$ 4.1}          & 70.5 $\pm$ 1.9 & 73.5 $\pm$ 4.3 & 73.2 $\pm$ 2.6 & 73.2 $\pm$ 4.1          & 70.1 $\pm$ 1.6          \\
                          & Attacked & 22.6 $\pm$ 1.7          & 26.2 $\pm$ 2.9          & 41.3 $\pm$ 1.9          & 32.1 $\pm$ 5.9 & 26.5 $\pm$ 5.9 & 21.1 $\pm$ 3.6 & 35.8 $\pm$ 3.2          & \textbf{43.1 $\pm$ 3.9} \\ \cline{2-10} 
\multirow{2}{*}{D\&D}     & Clean    & 73.5 $\pm$ 2.9 & 72.9 $\pm$ 3.1 & 73.5 $\pm$ 3.4 & 71.8 $\pm$ 0.9 & \textbf{74.4 $\pm$ 4.1} & 72.1 $\pm$ 4.9 & 73.8 $\pm$ 1.9 & 70.9 $\pm$ 3.6          \\
                          & Attacked & 12.9 $\pm$ 3.7 & 28.9 $\pm$ 5.6 & 5.6 $\pm$ 2.9  & 24.1 $\pm$ 4.9 & 23.9 $\pm$ 5.1          & 27.5 $\pm$ 3.6 & 32.9 $\pm$ 3.2 & \textbf{43.7 $\pm$ 5.1} \\ \cline{2-10}
\multirow{2}{*}{ENZYMES}  & Clean    & 33.9 $\pm$ 4.9          & 27.8 $\pm$ 0.8          & \textbf{37.2 $\pm$ 4.3} & 27.8 $\pm$ 4.1 & 34.4 $\pm$ 0.8 & 30.0 $\pm$ 3.6 & 23.9 $\pm$ 2.8          & 27.8 $\pm$ 4.1          \\
                          & Attacked & 1.1 $\pm$ 0.8           & 0.6 $\pm$ 0.8           & 2.8 $\pm$ 2.1           & 2.5 $\pm$ 1.2  & 2.8 $\pm$ 0.8  & 5.6 $\pm$ 1.2  & 5.6 $\pm$ 0.8           & \textbf{6.1 $\pm$ 2.1}  \\ \cline{2-10} 
\multirow{2}{*}{IMDB-B}   & Clean    & 64.3 $\pm$ 2.1          & 61.7 $\pm$ 4.7          & 52.3 $\pm$ 4.5          & 63.7 $\pm$ 4.1 & 61.0 $\pm$ 5.7 & 60.3 $\pm$ 4.7 & \textbf{66.0 $\pm$ 3.8} & 63.3 $\pm$ 4.7          \\
                          & Attacked & 43.7 $\pm$ 4.2          & 40.0 $\pm$ 5.1          & 42.0 $\pm$ 4.5          & 38.3 $\pm$ 4.3 & 34.3 $\pm$ 4.3 & 43.9 $\pm$ 2.1 & 35.0 $\pm$ 4.6          & \textbf{51.0 $\pm$ 3.8} \\ \cline{2-10} 
\multirow{2}{*}{ER\_MD}   & Clean    & 68.2 $\pm$ 0.6          & \textbf{72.7 $\pm$ 2.8} & 71.2 $\pm$ 1.9          & 69.3 $\pm$ 4.6 & 67.8 $\pm$ 3.8 & 68.9 $\pm$ 2.8 & 61.8 $\pm$ 5.6          & 63.6 $\pm$ 2.4          \\
                          & Attacked & 34.1 $\pm$ 3.2          & 35.6 $\pm$ 2.9          & 34.1 $\pm$ 1.4          & 36.7 $\pm$ 3.2 & 36.7 $\pm$ 3.2 & 35.6 $\pm$ 1.4 & 40.8 $\pm$ 2.3          & \textbf{60.1 $\pm$ 2.8} \\ \cline{2-10} 
\multirow{2}{*}{MSRC\_9}  & Clean    & \textbf{93.9 $\pm$ 2.1} & 89.4 $\pm$ 2.1          & 93.2 $\pm$ 4.9          & 91.7 $\pm$ 1.1 & 93.2 $\pm$ 1.9 & 92.4 $\pm$ 1.1 & 91.7 $\pm$ 2.8          & 91.7 $\pm$ 4.6          \\
                          & Attacked & 65.9 $\pm$ 7.4          & 72.7 $\pm$ 3.7          & 60.6 $\pm$ 2.1          & 67.5 $\pm$ 3.8 & 69.6 $\pm$ 5.2 & 54.6 $\pm$ 3.7 & 65.2 $\pm$ 7.7          & \textbf{83.3 $\pm$ 4.2} \\ \hline
\end{tabular}}
\end{table}

%% file: Tables/3_defense_methods.tex
\begin{table}[]
\small

\caption{Clean and Attacked classification accuracy ($\pm$ standard deviation) of the RS-Pool and other pooling strategies applied on different graph adversarial methods when subject to the PGD adversarial attacks.}
\label{tab:results_defense}
\renewcommand{\arraystretch}{1.5}

\resizebox{\textwidth}{!}{%
\begin{tabular}{lllcccccccc}
\hline
 &
   &
   &
  Sum &
  Average &
  Max &
  SAG &
  TopK-P &
  PAN-P &
  Sort-P &
  RS-Pool \\ \hline
\multirow{8}{*}{\rotatebox{90}{PROTEINS}} &
  \multirow{2}{*}{\begin{tabular}[c]{@{}l@{}}Adv\\ Train.\end{tabular}} &
  Clean &
  72.4 $\pm$ 4.0 &
  70.2 $\pm$ 2.3 &
  73.2 $\pm$ 3.7 &
  71.1 $\pm$ 3.7 &
  73.7 $\pm$ 2.9 &
  73.5 $\pm$ 2.2 &
  \textbf{73.8 $\pm$ 2.3} &
  72.1 $\pm$ 3.1 \\
 &
   &
  Attacked &
  45.8 $\pm$ 5.1 &
  31.3 $\pm$ 2.2 &
  18.5 $\pm$ 3.7 &
  49.1 $\pm$ 3.6 &
  46.1 $\pm$ 4.8 &
  32.7 $\pm$ 3.1 &
  51.5 $\pm$ 3.5 &
  \textbf{54.8 $\pm$ 4.1} \\ \cline{3-11} 
 &
  \multirow{2}{*}{SoftMedian} &
  Clean &
  72.6 $\pm$ 4.7 &
  71.2 $\pm$ 1.4 &
  73.2 $\pm$ 4.6 &
  72.0 $\pm$ 0.9 &
  73.2 $\pm$ 0.8 &
  73.0 $\pm$ 5.9 &
  \textbf{73.7 $\pm$ 0.7} &
  74.3 $\pm$ 1.3 \\
 &
   &
  Attacked &
  53.9 $\pm$ 4.9 &
  42.8 $\pm$ 6.7 &
  41.5 $\pm$ 5.1 &
  36.3 $\pm$ 2.9 &
  52.4 $\pm$ 1.2 &
  41.7 $\pm$ 2.1 &
  58.3 $\pm$ 1.8 &
  \textbf{63.2 $\pm$ 2.8} \\ \cline{3-11} 
 &
  \multirow{2}{*}{RGCN} &
  Clean &
  72.6 $\pm$ 2.3 &
  61.0 $\pm$ 1.6 &
  57.6 $\pm$ 3.4 &
  66.7 $\pm$ 2.2 &
  70.2 $\pm$ 3.4 &
  70.5 $\pm$ 1.9 &
  70.8 $\pm$ 0.4 &
  71.7 $\pm$ 1.1 \\
 &
   &
  Attacked &
  65.4 $\pm$ 0.8 &
  58.6 $\pm$ 1.7 &
  39.8 $\pm$ 4.2 &
  59.2 $\pm$ 1.1 &
  63.1 $\pm$ 3.4 &
  61.3 $\pm$ 4.7 &
  63.6 $\pm$ 2.3 &
  67.3 $\pm$ 1.8 \\ \cline{3-11} 
 &
  \multirow{2}{*}{\begin{tabular}[c]{@{}l@{}}Randomized\\ Smoothing\end{tabular}} &
  Clean &
  75.4 $\pm$ 3.6 &
  71.7 $\pm$ 3.3 &
  72.6 $\pm$ 2.2 &
  70.8 $\pm$ 2.3 &
  74.1 $\pm$ 3.6 &
  75.6 $\pm$ 3.1 &
  74.4 $\pm$ 4.1 &
  73.5 $\pm$ 3.4 \\
 &
   &
  Attacked &
  56.2 $\pm$ 1.5 &
  49.7 $\pm$ 3.7 &
  36.3 $\pm$ 3.6 &
  57.2 $\pm$ 4.4 &
  56.1 $\pm$ 3.2 &
  53.6 $\pm$ 0.7 &
  52.9 $\pm$ 4.3 &
  59.3 $\pm$ 4.8 \\ \cline{2-11} 
\multirow{8}{*}{\rotatebox{90}{D\&D}} &
  \multirow{2}{*}{\begin{tabular}[c]{@{}l@{}}Adv\\ Train.\end{tabular}} &
  Clean &
  \textbf{76.7 $\pm$ 3.4} &
  68.9 $\pm$ 4.1 &
  70.4 $\pm$ 4.3 &
  69.3 $\pm$ 3.6 &
  75.1 $\pm$ 4.3 &
  74.9 $\pm$ 2.6 &
  76.0 $\pm$ 2.5 &
  74.7 $\pm$ 3.4 \\
 &
   &
  Attacked &
  7.3 $\pm$ 2.7 &
  18.6 $\pm$ 6.3 &
  6.7 $\pm$ 2.2 &
  19.6 $\pm$ 5.1 &
  8.7 $\pm$ 1.7 &
  7.1 $\pm$ 2.6 &
  7.3 $\pm$ 1.5 &
  \textbf{31.9 $\pm$ 4.9} \\ \cline{3-11} 
 &
  \multirow{2}{*}{SoftMedian} &
  Clean &
  74.1 $\pm$ 3.4 &
  68.9 $\pm$ 4.7 &
  69.0 $\pm$ 5.1 &
  68.4 $\pm$ 5.1 &
  75.7 $\pm$ 5.8 &
  \textbf{77.3 $\pm$ 1.6} &
  74.9 $\pm$ 2.9 &
  72.9 $\pm$ 3.4 \\
 &
   &
  Attacked &
  14.9 $\pm$ 3.8 &
  51.2 $\pm$ 4.3 &
  8.5 $\pm$ 3.8 &
  46.1 $\pm$ 5.6 &
  10.9 $\pm$ 4.3 &
  29.1 $\pm$ 6.4 &
  10.4 $\pm$ 4.4 &
  \textbf{54.3 $\pm$ 3.8} \\ \cline{3-11} 
 &
  \multirow{2}{*}{RGCN} &
  Clean &
  74.9 $\pm$ 1.3 &
  63.8 $\pm$ 2.4 &
  66.4 $\pm$ 5.1 &
  70.6 $\pm$ 3.4 &
  75.4 $\pm$ 2.1 &
  \textbf{76.8 $\pm$ 1.8} &
  73.1 $\pm$ 2.4 &
  71.2 $\pm$ 2.8 \\
 &
   &
  Attacked &
  37.8 $\pm$ 2.1 &
  22.8 $\pm$ 3.6 &
  13.4 $\pm$ 2.9 &
  38.7 $\pm$ 4.8 &
  44.5 $\pm$ 2.4 &
  58.7 $\pm$ 3.4 &
  53.7 $\pm$ 3.6 &
  \textbf{61.4 $\pm$ 3.2} \\ \cline{3-11} 
 &
  \multirow{2}{*}{\begin{tabular}[c]{@{}l@{}}Randomized\\ Smoothing\end{tabular}} &
  Clean &
  74.9 $\pm$ 4.3 &
  72.1 $\pm$ 2.8 &
  42.5 $\pm$ 2.7 &
  69.3 $\pm$ 1.7 &
  74.8 $\pm$ 1.5 &
  \textbf{76.4 $\pm$ 1.5} &
  76.1 $\pm$ 1.9 &
  72.1 $\pm$ 3.8 \\
 &
   &
  Attacked &
  39.9 $\pm$ 2.6 &
  43.1 $\pm$ 3.5 &
  22.8 $\pm$ 4.5 &
  34.1 $\pm$ 4.1 &
  45.1 $\pm$ 4.8 &
  41.1 $\pm$ 3.5 &
  42.2 $\pm$ 1.7 &
  \textbf{49.7 $\pm$ 4.6} \\ \cline{2-11} 
\multirow{8}{*}{\rotatebox{90}{ER\_MD}} &
  \multirow{2}{*}{\begin{tabular}[c]{@{}l@{}}Adv\\ Train.\end{tabular}} &
  Clean &
  63.7 $\pm$ 2.3 &
  63.3 $\pm$ 3.4 &
  63.3 $\pm$ 4.6 &
  62.2 $\pm$ 4.5 &
  63.3 $\pm$ 4.1 &
  \textbf{68.1 $\pm$ 0.9} &
  65.5 $\pm$ 4.8 &
  63.3 $\pm$ 2.8 \\
 &
   &
  Attacked &
  42.6 $\pm$ 4.1 &
  45.7 $\pm$ 2.9 &
  44.2 $\pm$ 2.6 &
  48.6 $\pm$ 1.4 &
  42.3 $\pm$ 4.9 &
  35.4 $\pm$ 4.2 &
  34.9 $\pm$ 3.3 &
  \textbf{53.9 $\pm$ 2.6} \\ \cline{3-11} 
 &
  \multirow{2}{*}{SoftMedian} &
  Clean &
  \textbf{70.0 $\pm$ 1.9} &
  62.9 $\pm$ 6.9 &
  62.5 $\pm$ 4.5 &
  62.5 $\pm$ 4.0 &
  65.1 $\pm$ 4.8 &
  66.7 $\pm$ 6.8 &
  65.5 $\pm$ 6.1 &
  62.5 $\pm$ 1.9 \\
 &
   &
  Attacked &
  36.3 $\pm$ 5.5 &
  42.3 $\pm$ 5.0 &
  49.1 $\pm$ 5.0 &
  43.4 $\pm$ 4.5 &
  42.3 $\pm$ 5.0 &
  26.6 $\pm$ 3.4 &
  34.5 $\pm$ 1.0 &
  \textbf{53.6 $\pm$ 3.4} \\ \cline{3-11} 
 &
  \multirow{2}{*}{RGCN} &
  Clean &
  58.0 $\pm$ 5.5 &
  60.1 $\pm$ 4.3 &
  58.4 $\pm$ 4.7 &
  62.9 $\pm$ 5.7 &
  63.6 $\pm$ 4.6 &
  \textbf{68.5 $\pm$ 6.4} &
  66.7 $\pm$ 3.7 &
  59.9 $\pm$ 3.7 \\
 &
   &
  Attacked &
  39.2 $\pm$ 1.2 &
  47.5 $\pm$ 2.1 &
  44.5 $\pm$ 3.9 &
  49.3 $\pm$ 5.1 &
  28.0 $\pm$ 3.8 &
  38.6 $\pm$ 1.4 &
  46.8 $\pm$ 1.4 &
  \textbf{51.7 $\pm$ 0.9} \\ \cline{3-11} 
 &
  \multirow{2}{*}{\begin{tabular}[c]{@{}l@{}}Randomized\\ Smoothing\end{tabular}} &
  Clean &
  \textbf{69.2 $\pm$ 1.0} &
  61.4 $\pm$ 3.4 &
  61.4 $\pm$ 4.4 &
  61.0 $\pm$ 3.5 &
  62.9 $\pm$ 4.2 &
  67.8 $\pm$ 2.3 &
  63.6 $\pm$ 5.4 &
  61.7 $\pm$ 4.1 \\
 &
   &
  Attacked &
  40.4 $\pm$ 3.4 &
  40.4 $\pm$ 6.4 &
  46.4 $\pm$ 1.1 &
  48.3 $\pm$ 3.4 &
  40.1 $\pm$ 4.7 &
  43.4 $\pm$ 5.1 &
  43.8 $\pm$ 4.4 &
  \textbf{54.7 $\pm$ 4.4} \\ \hline
\end{tabular}%
}
\end{table}

%% file: Tables/5_bit_flip_attacks.tex
\begin{table}[]
\caption{Attacked classification accuracy ($\pm$ standard deviation) of the RS-Pool and other pooling strategies on different graph classification dataset when subject to Bit-Flip Attacks.}
\label{tab:bitflip_results}
\renewcommand{\arraystretch}{1.3}
\resizebox{\columnwidth}{!}{%
\begin{tabular}{lcccccccc}
\hline
Dataset  & Sum        & Average    & Max        & SAG        & TopK-P     & PAN-P      & Sort-P     & RS-Pool    \\ \hline
NCI1     & 64.7 $\pm$ 1.9 & 62.2 $\pm$ 1.9 & 65.9 $\pm$ 1.3 & 65.2 $\pm$ 1.3 & 61.2 $\pm$ 1.7 & 59.2 $\pm$ 1.7 & 63.5 $\pm$ 1.7 & \textbf{66.0 $\pm$ 1.6} \\
DD       & 59.8 $\pm$ 4.1 & 61.8 $\pm$ 2.7 & 63.6 $\pm$ 2.6 & 63.1 $\pm$ 1.4 & 58.4 $\pm$ 2.1 & 56.2 $\pm$ 1.5 & 56.8 $\pm$ 2.5 & \textbf{68.3 $\pm$ 1.0} \\
ER\_MD   & 55.2 $\pm$ 2.5 & 52.3 $\pm$ 1.8 & 51.1 $\pm$ 1.8 & 52.4 $\pm$ 2.0 & 52.9 $\pm$ 2.7 & 52.2 $\pm$ 1.1 & 51.9 $\pm$ 2.7 & \textbf{55.5 $\pm$ 2.5} \\
PROTEINS & 57.0 $\pm$ 5.8 & 64.2 $\pm$ 1.5 & 64.8 $\pm$ 2.7 & 63.7 $\pm$ 0.9 & 61.2 $\pm$ 5.0 & 60.0 $\pm$ 4.9 & 66.7 $\pm$ 4.7 & \textbf{69.6 $\pm$ 1.1} \\ \hline
\end{tabular}%
}
\end{table}

%% file: Appendix/6_Experimental_Details.tex
\subsection{Datasets}
For our experiments, we use standard graph classification datasets from the TUDataset benchmark~\cite{morris2020tudataset}, selecting a diverse subset to represent different domains. Specifically, we include datasets from bioinformatics and chemoinformatics (PROTEINS, ENZYMES, D\&D), small molecule prediction (NCI1, ER\_MD), and image-based graphs (MSRC\_9). Model evaluation follows the standardized protocol of \citet{errica2020fair}, using 10-fold cross-validation. When public folds are available, we use them directly; otherwise, we generate new folds following the same procedure. Dataset statistics and characteristics are summarized in Table~\ref{tab:data_statistics}.

\begin{table}[h]
\caption{Statistics of the graph classification datasets used in our experiments.}
\label{tab:data_statistics}
\begin{center}
\begin{small}
\begin{sc}
\begin{tabular}{lcccc}
\toprule
Dataset & \#Graphs & \#Nodes & \#Edges & \#Classes \\
\midrule
PROTEINS    & 1113 & 39.06 & 72.82 & 2 \\
DD    & 1178 & 284.32 & 715.66 & 2 \\
NCI1    & 4110 & 29.87 & 32.30 & 2 \\
ENZYMES    & 600 & 32.63 & 62.14 & 6 \\
IMDB-B    & 1000 & 19.77 & 96.53 & 2 \\
REDDIT-B    & 2000 & 429.63 & 497.75 & 2 \\
ER\_MD    & 446 & 21.33 & 234.85 & 2 \\
MSRC\_9    & 221 & 40.58 & 97.94 & 8\\
\bottomrule
\end{tabular}
\end{sc}
\end{small}
\end{center}
\end{table}

\subsection{Experimental Setup}
In all of the experiments, we use a 2-Layers GCN model. Specifically, the model uses a 2-layer convolutional architecture (consisting of two iterations of message passing and updating) stacked with a Multi-Layer Perception (MLP) as a readout. The only change in the architecture was on the pooling level since the intent was to compare the different benchmarks in an iso-architectural setting, to ensure a fair evaluation of their robustness. 

All models are trained using the Adam optimizer~\citep{kingma2014adam} with a learning rate of $1 \times 10^{-3}$ for 100 epochs. We set the hidden feature dimension to 32 and we used ReLU as our activation function for all the models. For all the experiments, we used the same initialization distribution and the same number of training epochs to ensure fairness following insights from previous work\cite{ennadir2024if}. Additionally, to account for variability due to random initialization, each experiment is repeated 10 times, and we report the mean and standard deviation of the results. The experiments have been run on a NVIDIA A40 GPU and we estimate the total number of hours of computing to be around 200 hours.

For our proposed RS-Pool, we set the number of iterations in the estimation algorithm to $K=2$ in the case of GCN while we set it to $K=5$ in the case of GIN. In addition, the temperature parameter $\tau$ was tuned separately for each dataset to optimize the trade-off between clean accuracy and adversarial robustness, as detailed in Section~\ref{sec:exp_results} (Figures~\ref{fig:analysis_of_risk} (e)). 

\subsection{Implementation Details}
Our implementation is provided in the supplementary materials and will be made publicly available after the review period. The code is developed using PyTorch~\citep{paszke2019pytorch}, with a dense implementation of GNNs, which is required for executing the considered adversarial attacks. For all baseline pooling methods, we adapt the official implementations from \textit{PyTorch Geometric} (PyG)~\citep{Fey/Lenssen/2019}, released under the MIT license.

%% file: main_paper/check_list.tex
\begin{enumerate}

\item {\bf Claims}
    \item[] Question: Do the main claims made in the abstract and introduction accurately reflect the paper's contributions and scope?
    \item[] Answer: \answerYes{} 
    \item[] Justification: The main claims presented in the abstract and introduction are supported by the theoretical results in Section~\ref{sec:robustness_analysis} and further validated through empirical evidence in Section~\ref{sec:exp_results}.

\item {\bf Limitations}
    \item[] Question: Does the paper discuss the limitations of the work performed by the authors?
    \item[] Answer: \answerYes{} 
    \item[] Justification: We discuss the main limitations of our method in Section~\ref{sec:conclusion}.

\item {\bf Theory assumptions and proofs}
    \item[] Question: For each theoretical result, does the paper provide the full set of assumptions and a complete (and correct) proof?
    \item[] Answer: \answerYes{}{} 
    \item[] Justification: All theoretical claims are supported by detailed proofs in Appendices~\ref{appendix:theo_gcn}, \ref{appendix:theo_gin}, \ref{appendix:theo_r_pool}, and~\ref{appendix:proof_remark}. Additionally, the problem setup and underlying assumptions are clearly defined in Section~\ref{sec:preliminaries} and are explicitly referenced in each theorem and proof.

    \item {\bf Experimental result reproducibility}
    \item[] Question: Does the paper fully disclose all the information needed to reproduce the main experimental results of the paper to the extent that it affects the main claims and/or conclusions of the paper (regardless of whether the code and data are provided or not)?
    \item[] Answer: \answerYes{} 
    \item[] Justification: In addition to presenting a detailed experimental setup in Appendix~\ref{appendix:experimental_details}, we include the source code for reproducing our results in the supplementary materials.

\item {\bf Open access to data and code}
    \item[] Question: Does the paper provide open access to the data and code, with sufficient instructions to faithfully reproduce the main experimental results, as described in supplemental material?
    \item[] Answer: \answerYes{} 
    \item[] Justification: All experiments were conducted using publicly available datasets. When available, we used the available public train/val/test folds, and we provide all of those. To ensure reproducibility, we provide the source code in the supplementary materials.

\item {\bf Experimental setting/details}
    \item[] Question: Does the paper specify all the training and test details (e.g., data splits, hyperparameters, how they were chosen, type of optimizer, etc.) necessary to understand the results?
    \item[] Answer: \answerYes{} 
    \item[] Justification: Appendix~\ref{appendix:experimental_details} provides details of the experimental setup, including details about the train/val/test used folds and the values of all hyperparameters.

\item {\bf Experiment statistical significance}
    \item[] Question: Does the paper report error bars suitably and correctly defined or other appropriate information about the statistical significance of the experiments?
    \item[] Answer: \answerYes{} 
    \item[] Justification: We report the mean and the standard deviation of all results in the respective tables, based on repeated experiments using 10 different random seeds. Further details are provided in Appendix~\ref{appendix:experimental_details}.

\item {\bf Experiments compute resources}
    \item[] Question: For each experiment, does the paper provide sufficient information on the computer resources (type of compute workers, memory, time of execution) needed to reproduce the experiments?
    \item[] Answer: \answerYes{} 
    \item[] Justification: We provide information about compute resources for our experiments in Appendix~\ref{appendix:experimental_details}.
    
\item {\bf Code of ethics}
    \item[] Question: Does the research conducted in the paper conform, in every respect, with the NeurIPS Code of Ethics \url{https://neurips.cc/public/EthicsGuidelines}?
    \item[] Answer: \answerYes{} 
    \item[] Justification:  Our work adheres to the NeurIPS Code of Ethics. We use public datasets and publicly available models and report on the limitations of our work.

\item {\bf Broader impacts}
    \item[] Question: Does the paper discuss both potential positive societal impacts and negative societal impacts of the work performed?
    \item[] Answer:  \answerNA{} 
    \item[] Justification: Our work has no further societal impacts apart from known impacts of graph-based models. We are not aware of any applications of our approach that would result in negative societal impacts.
    
\item {\bf Safeguards}
    \item[] Question: Does the paper describe safeguards that have been put in place for responsible release of data or models that have a high risk for misuse (e.g., pretrained language models, image generators, or scraped datasets)?
    \item[] Answer: \answerNA{} 
    \item[] Justification: Our paper does not release any data or models that have a high risk for misuse.

\item {\bf Licenses for existing assets}
    \item[] Question: Are the creators or original owners of assets (e.g., code, data, models), used in the paper, properly credited and are the license and terms of use explicitly mentioned and properly respected?
    \item[] Answer: \answerYes{} 
    \item[] Justification: All datasets, models and repositories were cited appropriately.

\item {\bf New assets}
    \item[] Question: Are new assets introduced in the paper well documented and is the documentation provided alongside the assets?
    \item[] Answer: \answerNA{} 
    \item[] Justification: Our paper does not release new assets, only rely on public datasets and models.

\item {\bf Crowdsourcing and research with human subjects}
    \item[] Question: For crowdsourcing experiments and research with human subjects, does the paper include the full text of instructions given to participants and screenshots, if applicable, as well as details about compensation (if any)? 
    \item[] Answer: \answerNA{} 
    \item[] Justification: Justification: Our work did not involve crowdsourcing nor research with human subjects. All experiments are performed on publicly available datasets.

\item {\bf Institutional review board (IRB) approvals or equivalent for research with human subjects}
    \item[] Question: Does the paper describe potential risks incurred by study participants, whether such risks were disclosed to the subjects, and whether Institutional Review Board (IRB) approvals (or an equivalent approval/review based on the requirements of your country or institution) were obtained?
    \item[] Answer: \answerNA{} 
    \item[] Justification: Our paper does not involve crowdsourcing nor research with human subjects, all experiments are performed on publicly available datasets.

\item {\bf Declaration of LLM usage}
    \item[] Question: Does the paper describe the usage of LLMs if it is an important, original, or non-standard component of the core methods in this research? Note that if the LLM is used only for writing, editing, or formatting purposes and does not impact the core methodology, scientific rigorousness, or originality of the research, declaration is not required.
    \item[] Answer: \answerNA{} 
    \item[] Justification: No LLM was used in this work for the core methods.

\end{enumerate}